\documentclass{article}
\usepackage{fullpage}
\usepackage[round]{natbib}

\usepackage{array}
\usepackage{multirow}
\usepackage{listings}
\usepackage{xspace}
\usepackage{xcolor}
\usepackage{hyperref}
\usepackage{url}
\usepackage{amsfonts}
\usepackage{amssymb}
\usepackage{amsmath}
\usepackage{amsthm}
\usepackage{graphicx}
\usepackage{booktabs}
\usepackage{placeins}
\usepackage{subcaption}
\usepackage{threeparttable}
\usepackage{enumitem}
\setlist{nosep,leftmargin=1.5em}

\usepackage{soul}

\definecolor{algbg}{RGB}{245,245,245}
\definecolor{algrule}{RGB}{0,0,0}
\definecolor{algkw}{RGB}{0,0,0}
\definecolor{algcomment}{RGB}{110,120,135}
\lstdefinestyle{sopqalg}{
  basicstyle=\fontencoding{T1}\fontfamily{zi4}\selectfont\footnotesize,
  columns=fixed,
  basewidth=0.44em,
  lineskip=1pt,
  keepspaces=true,
  upquote=true,
  breaklines=false,
  frame=leftline,
  framerule=2pt,
  rulecolor=\color{algrule},
  backgroundcolor=\color{algbg},
  framesep=4pt,
  xleftmargin=4pt,
  aboveskip=4pt,
  belowskip=4pt,
  morekeywords={for,each,in,return,while},
  keywordstyle=\color{algkw}\bfseries,
  emph={fit,encode,reconstruct,score},
  emphstyle=\color{algrule}\bfseries,
  moredelim=[l][\color{algcomment}]{\#},
  literate={<}{{\textless}}1 {>}{{\textgreater}}1
           {<-}{{$\leftarrow$}}2 {<=}{{$\le$}}2 {*}{{$\ast$}}1 {-}{{-}}1
}

\soulregister{\ref}{7} 

\newcommand{\code}[1]{\texttt{#1}}
\newcommand{\vqb}{VQ-bench\xspace}

\newcommand{\alg}{\ifmmode \mathrm{ALG} \else ALG\xspace \fi} 
\newcommand{\ezpq}{\ifmmode \mathrm{EZPQ} \else EZPQ\xspace \fi} 
\newcommand{\sopq}{\ifmmode \mathrm{EZPQ} \else EZPQ\xspace \fi} 
\newcommand{\rwf}{\ifmmode \mathrm{RWF} \else RWF\xspace \fi}  
\newcommand{\zg}{\ifmmode \mathrm{ZG} \else ZG\xspace \fi} 
\newcommand{\crwf}{\ifmmode \mathrm{CRWF} \else CRWF\xspace \fi} 
\newcommand{\ez}{\ifmmode \mathrm{EZ} \else EZ\xspace \fi} 
\newcommand{\lb}{\ifmmode \mathrm{LB} \else LB\xspace \fi} 

\makeatletter
\renewcommand{\paragraph}{\@startsection{paragraph}{4}{\z@}{0.9ex plus
0.3ex minus .2ex}{-0.8em}{\normalsize\itshape}}
\makeatother

\newcommand{\R}{\mathbb{R}}
\newcommand{\E}{\mathbb{E}}

\newcommand{\tr}{\mathrm{Tr}}
\newcommand{\diag}{\mathrm{diag}}

\newtheorem{theorem}{Theorem}

\newtheorem{lemma}[theorem]{Lemma}

\theoremstyle{definition}

\title{Provably Near-Optimal PQ in the Extended Zador Model}

\author{
Edo Liberty \\
Pinecone \& Google \\
\texttt{edo@google.com}
\and
Ashwin Padaki\thanks{Supported by the National Science Foundation (NSF) GRFP under Grant No. DGE-2236662, and Grant No. CCF-2337993.} \\
UPenn \& Pinecone \\
\texttt{apadaki@engineering.upenn.edu}
\and
Amir Ingber \\
Pinecone \\
\texttt{ingber@pinecone.io}
}
\date{}

\begin{document}
\maketitle

\begin{abstract}
This paper investigates Product Quantization (PQ) Optimization with Gaussian data.
It offers three key contributions
(i) a proof that SVD is the optimal preconditioning rotation for PQ;
(ii) a new Extended Zador (\ez) estimator of the $k$-means quantization cost that is accurate in low bit rates and low dimensions;
and
(iii) a novel segmentation algorithm based on 1-d bin packing of global bit allocations. 
The result is a simple algorithm (\ezpq) with strong approximation guarantees.
It matches OPQ's empirical accuracy while providing a per-instance distortion lower bound certificate under the Gaussian \ez cost model.
On typical real embedding spectra, \ezpq with $2$ bits per coordinate is often certifiably within $1\%$ of the optimum in this model. 
It also consistently achieves a lower reconstruction error than all random-rotation-based algorithms such as EDEN, E-RaBitQ, and TurboQuant.
\end{abstract}

\section{Introduction}
\label{sec:intro}

Vector Quantization (VQ)~\citep{shannon1948,shannon1959,Gersho1991VectorQA} is a cornerstone of modern AI infrastructure.
Vector databases (e.g.,\ Pinecone~\citep{pinecone}) rely on Product Quantization (PQ), Optimized Product Quantization (OPQ), and score-aware Anisotropic VQ~\citep{jegou2011pq,ge2013opq,ge2014opqJournal,johnson2017faiss,guo2020scann,simhadri2022bigann} for billion-scale approximate nearest neighbor (ANN) search, while large language models (LLMs) quantize model weights~\citep{frantar2023optq,lin2023awq,dettmers2023qlora,chee2023quip,tseng2024quipsharp} and key-value (KV) caches~\citep{liu2024kivi,hooper2024kvquant,kang2024gear} down to between $0.5$ and $4$ bits per coordinate; see~\citep{padaki2026vqbench} for a comprehensive survey and comparative benchmarks.

Recent fast quantizers~\citep{vargaftik2022eden,gao2024rabitq,gao2024erabitq,zandieh2025turboquant,ashkboos2024quarot} precondition data with a random projection or a random rotation~\citep{jl1984,ailon2009srht,ailon2009dualbch} so that all coordinates share identical marginal variance and a single scalar codebook.
This is optimal \emph{only} if every coordinate is artificially forced to use the exact same bit budget.
These quantizers are model free and extremely fast but, in general, do not give the highest compression rates.

State-of-the-art compression is achieved by data-dependent learned quantizers.
Classical Transform Coding~\citep{huang1963,segall1976,brandt2010transform} decorrelates coordinates via PCA/SVD and allocates variable bits to scalar quantizers.
It, however, forfeits multi-dimensional space-filling gains achievable by vector quantization.
Product Quantization (PQ)~\citep{jegou2011pq} uses $k$-means in segments but uses no preconditioning rotation or segment allocation. 
OPQ~\citep{ge2013opq,ge2014opqJournal} combines the two and is repeatedly reported by independent studies as the state-of-the-art quantization method~\citep{paparrizos2022vaq}.
Non-Parametric OPQ alternates between $d \times d$ Procrustes SVD updates and $k$-means re-clustering. 
Parametric OPQ uses one-shot SVD preconditioning and then greedily allocates coordinates to equal-width PQ segments based on their variance.
Alternative PCA-based methods such as Variance-Aware Quantization (VAQ)~\citep{paparrizos2022vaq} vary both the subspace widths and the dictionary sizes $k_r$. 
Parametric OPQ stands out because it is both simple to implement and gives very strong empirical results. 
Our analysis corroborates that it is, in fact, almost optimal in many settings.

In this paper, we formulate the exact PQ optimization problem for Gaussian data $X \sim \mathcal{N}(\mu, \Sigma)$.
We start by proving that the optimal preconditioning rotation for exact Gaussian $k$-means PQ is the SVD basis $U^\top$ (Section~\ref{sec:rotation}); conversely, random rotations approximate the worst preconditioner (Appendix~\ref{app:random-rotation}).
Next, we introduce the \emph{Extended Zador} (\ez) estimator of the $k$-means cost of non-isotropic Gaussian segments (Section~\ref{sec:ezg}), which proposes an empirical unification of the two asymptotic regimes - large dimensions (Shannon's rate-distortion theory), and 
the standard Zador-Gersho which is inaccurate for low rates and skewed spectra (Appendix~\ref{app:coupled-zador-validation}). 
Under the \ez cost model, we derive a global equal-slope bit allocation that yields an instance-specific lower bound on the quantization error of \emph{any} partition (Sections~\ref{sec:crwf}-\ref{sec:crwf-alloc}).
Finally, 1-d bin packing of these bit allocations provides the \ezpq segmentation (Section~\ref{sec:algorithm}). 
Unlike the OPQ variant, we do not restrict the PQ segments to be of the same size. 
While bin packing is itself computationally hard, it is a well understood problem with highly effective approximation algorithms.
We prove that EZPQ’s quantization error has only a second-order dependence on the packing discrepancies.
We empirically show that on real embedding spectra, \ezpq is typically within $1\%$ of the lower bound at up to $2$ bits per coordinate, and within $3\%$ at $4$ bits (Section~\ref{sec:results-certificates}).

\subsection{Problem Definition and Setup}
\label{sec:setup}

Define the PQ optimization problem as follows. Take as input a data matrix $X \in \R^{d \times n}$, a number $k = 2^C$ of $k$-means centers per segment (i.e., a budget of $C$ bits per segment), and a number of segments $m$. Produce a rotation matrix $R \in O(d)$ and $m$ disjoint segments $G = \{g_1, \dots, g_m\}$ with $g_r \subseteq [d]$ of widths $d_r = |g_r|$ (we also write $d_g = |g|$ for a generic segment $g$).
The cost is measured by the average per-vector $k$-means cost over $m$ different segments (rotated and centered).
Coordinates that are not a part of any segment incur the full cost of being un-quantized.   
More explicitly, let $\tilde{X} \;=\; R(X - \mu_X)$ (subtracting the empirical mean $\mu_X \in \R^d$ column-wise). Let $g_0 = [d] \setminus \bigcup_{r=1}^{m} g_r$ be the un-quantized set. Let $\tilde{X}_g$ denote $\tilde{X}$ restricted to the rows in $g$. Then, 
$\mathrm{cost}(R,G) = \frac{1}{n}\sum_{r=1}^{m} \operatorname{kmeans-cost}(\tilde{X}_{g_r}, k) + \frac{1}{n}\|\tilde{X}_{g_0}\|_F^2$. The goal of PQ optimization is to produce $R$ and $G$ that minimize $\mathrm{cost}(R,G)$. 
Throughout, $\mathrm{OPT} \;=\; \min_{R \in O(d)}\;\min_{G}\; \mathrm{cost}(R, G).$

\paragraph{Random Gaussian data:} This paper models data as $d$-dimensional Gaussian random vectors $X \sim \mathcal{N}(\mu, \Sigma)$ with mean $\mu \in \R^d$ and covariance $\Sigma \succ 0$. This assumption is commonly made (e.g.,~\citep{ge2013opq,ge2014opqJournal}) because, among data with covariance $\Sigma$, random Gaussians are the hardest to approximate in the rate distortion sense~\citep{cover2006it}, that is, under optimal coding of long blocks. This does not extend to $k$-means at a fixed rate with a codebook fit to finite data.
After centering $X - \mu$ and applying an orthogonal preconditioning rotation $R \in O(d)$, grouping is performed on vectors
$\tilde{X} \;=\; R(X - \mu) \;\sim\; \mathcal{N}(0, \tilde{\Sigma})$ where $\tilde{\Sigma} \;=\; R\,\Sigma\,R^\top$.

\paragraph{Gaussian $k$-means objective:}
For a segment $g$, let $\tilde{\Sigma}_g$ denote the principal submatrix of $\tilde{\Sigma}$ restricted to $g$, and let $e^2(\tilde{\Sigma}_g)$ denote the optimal expected \emph{per-coordinate} squared error of $k$-means for vectors $\tilde{x} \sim \mathcal{N}(0, \tilde{\Sigma}_g)$.
The global objective over rotations $R \in O(d)$ and partitions $G$ is
\begin{equation}
\label{eq:global-pq-obj}
\mathrm{OPT} \;=\; \min_{R \in O(d)}\;\min_{G}\; \mathrm{cost}(R, G),
\qquad
\mathrm{cost}(R, G) \;=\; \sum_{g \in G} d_g\, e^2(\tilde{\Sigma}_g) + \tr(\tilde{\Sigma}_{g_0}),
\qquad
\tilde{\Sigma} = R\,\Sigma\,R^\top .
\end{equation}
For the rest of the paper we use the shorthand notation $e^2(g)$ for $e^2(\tilde{\Sigma}_g)$. From here on, $\mathrm{cost}(R, G)$ and $\mathrm{OPT}$ refer to this population (Gaussian) objective, not to the empirical one defined above.

\section{SVD is the Optimal Preconditioning}
\label{sec:rotation}

We begin by proving that rotating the data by its left singular vectors (the eigenvectors of its covariance matrix) and making the covariance matrix diagonal is always the optimal rotation for the exact Gaussian $k$-means objective~\eqref{eq:global-pq-obj}.
Intuitively, the segment distortion is concave in the diagonal entries of $\tilde{\Sigma}$ while orthogonal rotations keep its diagonal entries inside a convex majorization polytope, so the minimum is attained at an extreme point corresponding (up to permutation) to the eigenvalue vector $\sigma^2$ achieved by the SVD rotation $R = U^\top$.

\begin{theorem}[Optimality of the SVD Rotation]
\label{thm:pca}
Let $U$ be the eigenbasis of $\Sigma$.
For the exact $k$-means objective~\eqref{eq:global-pq-obj}, the SVD rotation $R = U^\top$ globally minimizes the PQ optimization cost over rotations $R$:
\[
\min_{R \in O(d)} \min_{G} \mathrm{cost}(R, G) = \min_{G} \mathrm{cost}(U^\top, G) \ .
\]
\end{theorem}

\begin{proof}
For any rotation $R \in O(d)$ let $\tilde{\Sigma} = R\,\Sigma\,R^\top$.
Since $k$-means is invariant to orthogonal rotations we may assume without loss of generality that $R$ diagonalizes each principal block $\tilde{\Sigma}_g$. Let $s_g^2 =\diag(\tilde{\Sigma}_g)$.
The total error then depends only on $s^2 = \diag(\tilde{\Sigma})$ as follows:
\begin{equation}
\label{eq:mse-s2-decomp}
\mathrm{cost}(R, G) \;=\; f_G(s^2) \;:=\; \sum_{g \in G} d_g\, e^2(\diag(s_g^2)) \;+\; \sum_{j \in g_0} s_j^2.
\end{equation}
By the Schur-Horn theorem~\citep{horn1954,marshall2011majorization}, $s^2 = \diag(R\Sigma R^\top)$ lies in the compact convex majorization polytope $\mathcal{M}(\sigma^2) = \mathrm{conv}\{P \sigma^2 : P \in \Pi_d\}$.

For exact $k$-means on a segment $g$ write $x_j = s_j z_j$ with $z_j \sim \mathcal{N}(0, 1)$. 
Map any quantizer $q$ of $x$ to a quantizer $Q$ for $z$ by $Q(z)_j = q(x)_j / s_j$.
This gives
\[
d_g\, e^2(g) \;=\; \inf_Q \sum_{j\in g} s_j^2\, \varepsilon_j^2(Q), \qquad \varepsilon_j^2(Q) := \E\big[(z_j - Q(z)_j)^2\big].
\]
For any \emph{fixed} quantizer $Q$ the sum is linear in $s_g^2$. That is, $\varepsilon^2(Q)$ is independent of $s_g^2$.
Since the pointwise infimum of linear functions is concave, $d_g\,e^2(\diag(s_g^2))$ is globally concave in $s_g^2$ (and permutation-symmetric, since $z$ is isotropic) on $\mathcal{M}(\sigma^2)$.

Because a concave function on the compact convex polytope $\mathcal{M}(\sigma^2) = \mathrm{conv}\{P \sigma^2 : P \in \Pi_d\}$ attains its minimum at an extreme point, $\min_{s^2 \in \mathcal{M}(\sigma^2)} f_G(s^2)$ is attained at $s^2 = P \sigma^2$ for some permutation matrix $P \in \Pi_d$, corresponding to $R = P U^\top$.
Since permuting coordinates by $P$ simply relabels the partition $G$ ($\mathrm{cost}(P U^\top, G) = \mathrm{cost}(U^\top, P^{-1}(G))$), minimizing over $G$ absorbs $P$.
\end{proof}

Theorem~\ref{thm:pca} extends a classical result of transform coding. 
For Gaussian sources with each transform coefficient coded by a scalar quantizer, the Karhunen-Lo\`eve transform (KLT), i.e., the SVD rotation $R = U^\top$, is optimal~\citep{goyal2000transform}. The proof of~\citet{goyal2000transform} uses the same majorization argument. For a fixed bit allocation, the scalar distortion has the form $\sum_j s_j^2 f(b_j)$ for a single function $f$, which is linear in $s^2$. Theorem~\ref{thm:pca} replaces scalar quantizers by $k$-means on segments of arbitrary widths and linearity by concavity.

Throughout the rest of the paper we work in the diagonal SVD basis ($R = U^\top$), where
$\tilde{X} \sim \mathcal{N}\big(0, \diag(\sigma_1^2, \dots, \sigma_d^2)\big)$, i.e., $s_j^2 = \sigma_j^2$.
The next section introduces the Extended Zador (\ez) model for $e^2(g)$.

\section{The Extended Zador (\ez) Model}
\label{sec:ezg}
This section presents the Extended Zador model for approximating $e^2(g)$ in the right regime for PQ.
PQ typically operates in the low bit-rate and low dimension setting. In this paper, we will use the convention of $k=256$ centroids per segment ($C=8$ bits).
The bit rate is usually set to the range $0.5$-$4$ and, as a result, the average segment dimension is usually in the range $2$-$16$. 

Shannon's distortion-rate function gives asymptotic estimators for $e^2(g)$ for any bit rate (Section~\ref{sec:rwf}) but underestimates at finite dimensions.
Zador and Gersho's result corrects for finite dimensions but is tight only in the asymptotically high bit rate regime (Section~\ref{sec:zg}).
 The Extended Zador formula (Section~\ref{sec:ez}), developed in this paper, combines the allocation structure of the former with the finite-dimension penalty of the latter to achieve a highly accurate estimate which is usually up to $2\%$ away from measured experimental values. 

\subsection{Shannon's Distortion-Rate Function and Reverse Water-Filling}
\label{sec:rwf}
Since we work in each segment's eigenbasis, coordinates in $g$ can be viewed as $d_g$ independent memoryless Gaussian sources (``channels''). 
Encoded jointly at a total rate of $C$ bits per time step, the optimal bit allocation per channel (coordinate) is given by Reverse Water-Filling~\citep{shannon1959,cover2006it}.
Let $D_j(b_j)$ denote the distortion of source $j$ with $b_j$ rate devoted to it. Then
\begin{equation}
\label{eq:rwf-program}
D_j(b_j) = \sigma_j^2\, 2^{-2b_j}
\qquad \text{and} \qquad 
d_g e_{\rwf}^2(g) \;:=\; \min_{b \ge 0}\; \sum_{j \in g} \sigma_j^2\, 2^{-2b_j}
\qquad \text{s.t.} \qquad 
\sum_{j \in g} b_j \;=\; C \ .
\end{equation}

The KKT conditions for minimizing $e_{\rwf}^2(g)$ equate $\sigma_j^2 2^{-2b_j} = \theta$ of all active ($b_j > 0$) coordinates.
We call $g$ admissible if all its coordinates are active ($\sigma_j^2 \ge \theta$).
Inactive coordinates with $\sigma_j^2 \le \theta$ lie below the water level, receive no bits, and incur their full variance as distortion.
Denoting $\bar{b}_g := C/d_g$, the \rwf bit allocation and error formula for admissible segments are:
\begin{equation}
\label{eq:rwf}
b_j \;=\; \frac{1}{2}\log_2 \frac{\sigma_j^2}{e_{\rwf}^2(g)} 
\qquad 
\text{and}
\qquad e_{\rwf}^2(g) = \Big(\prod_{j\in g}\sigma_j^2 \, 2^{-2b_j}\Big)^{1/d_g} = \big|\tilde{\Sigma}_g\big|^{1/d_g}\, 2^{-2\bar{b}_g} \ .
\end{equation}

Note that by the converse of the rate-distortion theorem $e_{\rwf}^2(g) \le e^2(g)$.
For PQ, this bound is tight only asymptotically for $d_g \rightarrow \infty$.
This causes \rwf to significantly underestimate the true $k$-means error in low dimensions. 
Section~\ref{sec:ezg-validation} shows experimentally that this causes up to $\sim 35\%$ underestimation.
The Zador-Gersho result quantifies and corrects for this effect, but only at high rate. 

\subsection{The Zador-Gersho High-Rate Result}
\label{sec:zg}
In finite dimensions, vector quantizers incur an additional space-filling inefficiency penalty.
Zador and Gersho's theorem ~\citep{zador1982,gersho1979,bucklew1982,gray1998} states the following.
For a fixed $d_g$, as $C \to \infty$, the per-coordinate distortion of the optimal $2^C$-point quantizer of $\mathcal{N}(0, \tilde{\Sigma}_g)$ tends to:

\begin{equation}
\label{eqn:zador}
e_{\mathrm{ZG}}^2(g) \;:=\; z(d_g)\big|\tilde{\Sigma}_g\big|^{1/d_g} 2^{-2\bar{b}_g}
\end{equation}

\begin{figure}[!htbp]
\centering
\includegraphics[width=\linewidth]{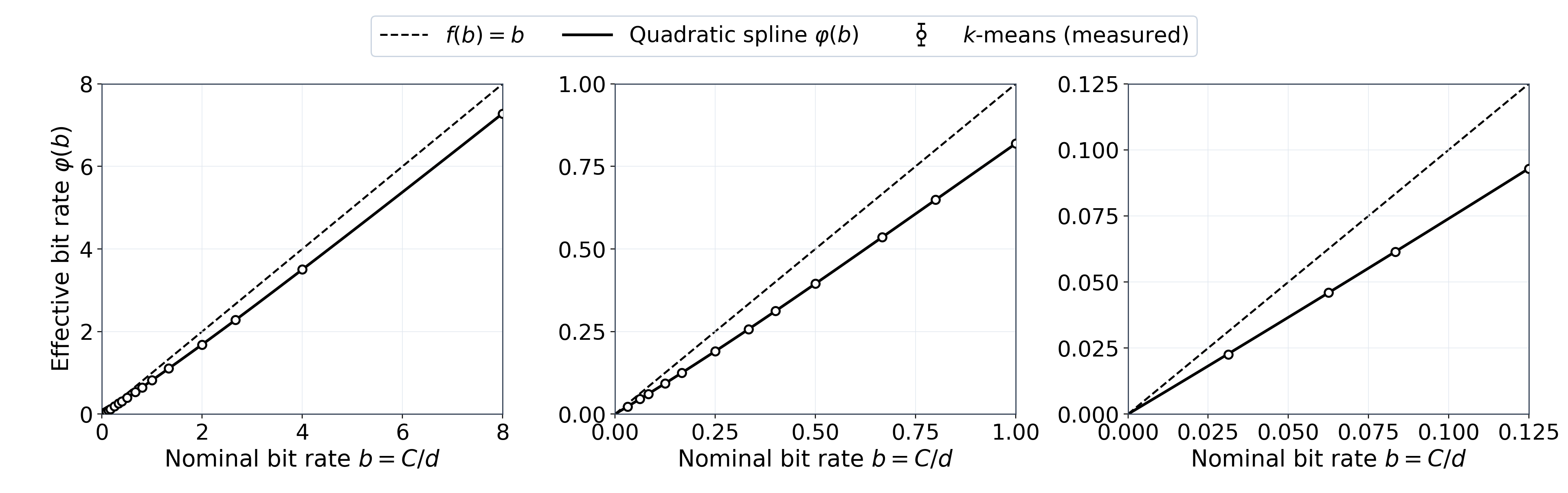}
\caption{Effective bit rate $\varphi(b)$ (for $C = 8$ bits) versus nominal bit rate $b = C/d$ on the ranges $b \le 8$ (left), $b \le 1$ (middle), and $b \le 0.125$ (right). The dashed line is the asymptotic rate $f(b) = b$, the solid line is the quadratic spline fit of $\varphi$ (Appendix~\ref{app:zador-approximations}), and the markers are empirical $k$-means measurements ($k = 256$; error bars over $5$ runs are smaller than the markers).}
\label{fig:phi-b}
\end{figure}

The function $z(d)$ in~\eqref{eqn:zador} is the classical space-filling constant $c_d$ normalized by the Gaussian factor, $z(d) = 2\pi\,(1 + 2/d)^{(d+2)/2}\, c_d$, which tends to $2\pi e\, c_d$ as $d \to \infty$.
Its largest value is $z(1) = \sqrt{3}\pi/2 \approx 2.72$ (the Panter-Dite constant) and it decreases with $d$ towards $\lim_{d\rightarrow \infty}z(d) = 1$ where Shannon's bound is recovered. It is instructive to view $z(d)$ as a rate penalty by defining the effective bit rate function $\varphi(b)$:
\begin{equation}
\label{eq:phi-def}
\varphi(b) \;:=\; b - \frac{1}{2}\log_2 z(C/b) \qquad \Longrightarrow \qquad e_{\mathrm{ZG}}^2(g) \;=\; \big|\tilde{\Sigma}_g\big|^{1/d_g}2^{-2\varphi(\bar{b}_g)}.
\end{equation}
The effective bit rate function $\varphi$ depends only on the budget $C$.\footnote{For simplicity, unless otherwise stated, $\varphi = \varphi_C$ for $C=8$.}
It does not depend on the segment's spectrum. See Figure~\ref{fig:phi-b}.

At the budgets where PQ operates, $k$-means departs from this result in two ways.
First, weak coordinates are effectively ignored by the clustering and incur their full variance as distortion, as in reverse water-filling (Section~\ref{sec:rwf}). For example, adding a coordinate with variance equal to $e^2(g)$ should not (and does not) increase $e^2(g)$.
Zador-Gersho has no such mechanism. Adding a coordinate at the implicit waterline $e_{\zg}^{2}$ seams to increase the predicted distortion. 
This effect has been studied, e.g., as phase transitions in deterministic annealing~\citep{rose1990statistical,rose1998deterministic}.
Second, the distortions of the active coordinates are not actually equal. 
In practice, coordinates with higher variance have \emph{lower} distortion! 
Both reverse water-filling and Zador-Gersho predict equal distortion for all active coordinates.
Other characterizations of the optimal distortion are asymptotic in the dimension~\citep{ingber2011dispersion,kostina2012fixed}, asymptotic in the rate, or scalar~\citep{max1960,marco2006low}. 
To our knowledge, none quantifies it at a fixed small dimension and a finite rate.
The \ez model (Section~\ref{sec:ez}) captures both effects (Theorem~\ref{thm:ez-limits}(v)-(vi)).

\subsection{The shape of the effective bit rate $\varphi$}
Zador-Gersho mathematically derive asymptotic values for $\varphi$ at $b = C/d$ for integer $d$.
Using a continuous and explicit function $\varphi$ has multiple benefits, chief among them is the ability to optimize smoothly over bit allocations.
In this paper we use a quadratic spline fitted to empirical values of $\varphi(b)$ (Appendix~\ref{app:zador-approximations}); a simple closed form works nearly as well.
Our proofs, however, do not rely on this specific choice of $\varphi$.
The claims in the paper hold for any $\varphi(b)$ that exhibits three properties on $[0, C]$:
\emph{(a) Origin limit:} $\varphi(0) := \lim_{b\rightarrow 0^+} \varphi(b) = 0$.
\emph{(b) Strict monotonicity and bounded slope:} $0 < \varphi'(b) \le 1$.
\emph{(c) Convexity and bounded curvature:} $0 \le \varphi''(b) < (2\ln 2)\varphi'(b)^2$. The only exceptions are the two limits of Theorem~\ref{thm:ez-limits}(ii) and (iv): they vary the budget $C$, and therefore need a family $\{\varphi_C\}$ and the additional conditions stated in Appendix~\ref{app:ez-limits}.

\subsection{The Extended Zador (\ez) Model}
\label{sec:ez}
The extended model uses both dynamic bit allocation like \rwf (which drops weak coordinates) and effective bit rates like Zador-Gersho (which account for finite dimensions).
In this model the distortion of coordinate $j$ is given by
\begin{equation}
\label{eq:ez}
D_j(b_j) \;:=\; \sigma_j^2\, 2^{-2\varphi(b_j)},
\end{equation}
where $\sum_{j\in g} b_j = C$ are implicit bit allocations.
Unlike Zador-Gersho, each coordinate receives its own effective bit rate $\varphi(b_j)$ instead of the segment's average 
$\varphi(\bar{b}_g)$.

\begin{equation}
\label{eq:seg-kkt-def}
d_g e_{\ez}^2(g) \;:=\; \min_{b \ge 0} \sum_{j \in g} \sigma_j^2\, 2^{-2\varphi(b_j)}
\qquad \text{subject to} \qquad
\sum_{j\in g} b_j \le C 
 \end{equation}
 
By the bounded curvature of $\varphi$ (property (c)), each $D_j$ is strictly convex. 
Therefore, the program is convex and its minimizer $b^{\ez}$ of~\eqref{eq:seg-kkt-def} equates the marginal slopes $D_j(b_j^{\ez})\,\varphi'(b_j^{\ez}) = \lambda_g$ across the interior active coordinates ($0< b_j^{\ez} < C$).

Appendix~\ref{app:ez-limits} shows that \ez behaves as intended (Theorem~\ref{thm:ez-limits}). It never falls below the Shannon bound ($e_{\rwf}^2(g) \le e_{\ez}^2(g)$), and, for $\varphi$ of the form~\eqref{eq:phi-def} (see Theorem~\ref{thm:ez-limits} for the precise conditions), it coincides with \zg on isotropic segments, recovers \zg as $C \to \infty$, and recovers \rwf as the dimension grows at a fixed rate per coordinate.
Within a segment, stronger active coordinates receive lower distortion (a property that we observed empirically, which is not captured by either the Shannon or the Zador theories), while coordinates with $\sigma_j^2 \le \theta_{\ez} := \lambda_g/\varphi'(0)$ receive no bits, as in \rwf.
Most importantly, \ez approximates the actual cost of $k$-means clustering on short segments with skewed spectra, at the operating budget $C$, much better than \rwf and \zg.

\subsection{Experimental Comparison of the Estimators}
\label{sec:ezg-validation}
\begin{figure}[!htbp]
\centering
\includegraphics[width=\linewidth]{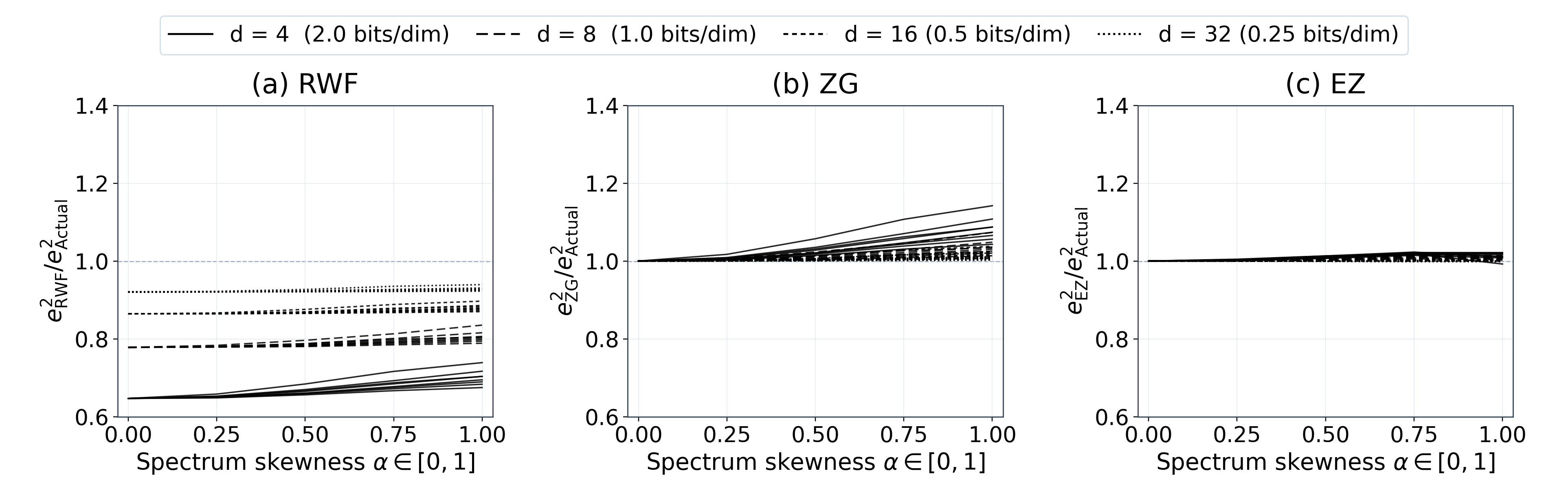}
\caption{Comparison of the three segment distortion estimators (a) \rwf~\eqref{eq:rwf}, (b) \zg~(\ref{eq:phi-def}), and (c) \ez~\eqref{eq:seg-kkt-def}, evaluated with the quadratic spline $\varphi(b)$ of Table~\ref{tab:quad-spline-knots} across $d \in \{4, 8, 16, 32\}$ (line dashing) and $8$ spectral shapes parameterized by skewness $\alpha \in \{0, 0.25, 0.5, 0.75, 1.0\}$. Actual $k$-means cost is the median over $5$ runs  with clusters trained on $1{,}000{,}000$ samples and cost evaluated on $200{,}000$ holdout samples with $k = 256$ (see Appendix~\ref{app:coupled-zador-validation}).}
\label{fig:crwf-validation}
\end{figure}

Figure~\ref{fig:crwf-validation} compares \rwf~\eqref{eq:rwf}, \zg~(\ref{eq:phi-def}), and \ez~\eqref{eq:seg-kkt-def} against empirical $k$-means distortion ($k = 256, C = 8$) across $d \in \{4, 8, 16, 32\}$ and $8$ spectral shapes parameterized by a skewness parameter $\alpha \in \{0, 0.25, 0.5, 0.75, 1.0\}$, ranging from isotropic ($\alpha = 0$) to maximally skewed ($\alpha = 1$).
\rwf underestimates distortion by $6\%$-$35\%$ ($18.41\%$ MAPE; Figure~\ref{fig:crwf-validation}a) because it ignores the space-filling penalty.
\zg is exact at $\alpha = 0$ (by construction, using empirical $\varphi$) but increasingly overestimates distortion as skewness grows, by up to $+14.22\%$ at $d = 4$ ($1.87\%$ MAPE; Figure~\ref{fig:crwf-validation}b).
\ez removes this skewness fan-out, reducing the non-isotropic error to $0.51\%$ MAPE (Figure~\ref{fig:crwf-validation}c).
Full experimental details and error statistics (Table~\ref{tab:crwf-validation}) are given in Appendix~\ref{app:coupled-zador-validation}.

For the remainder of the paper we adopt the \ez approximation $e^2(g) := e_{\ez}^2(g)$ in~\eqref{eq:global-pq-obj}. Accordingly, from here on $\mathrm{cost}$, $\mathrm{OPT}$, $\alg$ and $\lb$ are quantities of this \ez surrogate objective. They agree with the exact Gaussian $k$-means quantities only up to the estimation error of \ez (Appendix~\ref{app:coupled-zador-validation}), and they are not bounds on the empirical distortion of non-Gaussian data.
Even in the diagonal SVD basis, minimizing~\eqref{eq:global-pq-obj} over variable-width partitions $G$ is a combinatorial problem because the bit allocations within each segment $g_r$ are coupled by the budget $C$.
Parametric OPQ~\citep{ge2014opqJournal} simplifies this by fixing equal widths $d_r = d/m$ across all $d$ coordinates, whereas \ezpq refines this design by allowing variable segment widths and pruning zero-bit tail coordinates.

\section{\ezpq: Segmentation reduced to Bit Packing}
This section shows that a near-optimal segmentation can be obtained by a two-step process.
First, assign a bit budget $b_j$ to each of the $d$ coordinates.
Then, partition the coordinates into $m$ groups of roughly $C$ total bits each.
We show that the optimality gap of \ezpq is directly controlled by the tightness of this packing.

\subsection{A Lower Bound on $\mathrm{OPT}$}
\label{sec:crwf}
\label{sec:lower-bound}
\label{sec:per-coord-quantities}
\label{sec:lb-crwf}
\label{sec:key-lemma}

We begin with a global minimization of $\sum_{j=1}^d D_j(b_j)$ that provides target bit allocations $b_j^{\lb}$ and lower-bounds $\mathrm{OPT} = \min_{G} \mathrm{cost}(G)$, where $\mathrm{cost}(G) := \mathrm{cost}(U^\top, G)$.

\begin{lemma}[Universal Lower Bound]
\label{thm:lb}
\label{lem:segment-bit-alloc}
The minimal global sum of individual coordinate distortions lower-bounds the total distortion of the optimal partition.
Define $\lb$ as 
\begin{equation}
\label{eq:lb-prog}
\lb \;:=\; \min_{b} \;\sum_{j=1}^d \sigma_j^2 2^{-2\varphi(b_j)} \quad\text{s.t.}\quad \forall_j\;\;  b_j \in [0,C] \quad\text{and}\quad \sum_{j=1}^d b_j \;\le\; mC.
\end{equation}
Then $\mathrm{OPT} \;\ge\; \lb$ with $\mathrm{OPT}$ evaluated under the \ez model, $e^2 := e_{\ez}^2$.

\end{lemma}

\begin{proof}
Let $G = \{g_1, \dots, g_m\}$ be any partition with unquantized set $g_0$. Let $b^{\mathrm{ALG}}_j$ be the \ez bit allocation~\eqref{eq:seg-kkt-def} of coordinate $j$ within its segment in $G$, and $b^{\mathrm{ALG}}_j = 0$ for $j \in g_0$. Then
\[
\mathrm{cost}(G)  = \sum_{r=1}^m d_r\,e^2(g_r) + \sum_{j\in g_0} \sigma_j^2 = \sum_{j=1}^d D_j(b^{\mathrm{ALG}}_j) \ge \lb
\]
The last inequality holds because $b^{\mathrm{ALG}}_j \le C$ and $\sum_{r =1}^{m} \sum_{j \in g_r} b^{\mathrm{ALG}}_j \le mC$, so $b^{\mathrm{ALG}}$ is a feasible
solution of~\eqref{eq:lb-prog}, whose global minimum is $\lb$.
\end{proof}

\subsection{Solving for $\lb$}
\label{sec:crwf-alloc}

We now explain how to compute the bit allocations $b_j^{\lb}$ that attain $\lb$.
For contrast, classical \rwf equates the distortion across active coordinates; applied globally, this rule reads
\begin{equation}
\label{eq:equal-dist-crwf}
D_j(b_j) \;=\; \sigma_j^2\, 2^{-2\varphi(b_j)} \;=\; \lambda \quad (\forall\, j : \sigma_j^2 > \lambda),
\qquad
\sum_{j:\, \sigma_j^2 > \lambda} \varphi^{-1}\!\left(\frac{1}{2}\log_2 \frac{\sigma_j^2}{\lambda}\right) \;=\; mC.
\end{equation}
When minimizing the sum of coordinate distortions $\sum_j D_j(b_j)$, both globally in $\lb$~\eqref{eq:lb-prog} and within a single segment in $e_{\ez}^2(g)$~\eqref{eq:seg-kkt-def}, the non-linearity of the effective bit rate $\varphi(b)$ adds a marginal efficiency factor $\varphi'(b)$, and~\eqref{eq:equal-dist-crwf} is no longer optimal.

By Properties (b) and (c) of $\varphi$ in Section~\ref{sec:ezg}, each $D_j(b) = \sigma_j^2 2^{-2\varphi(b)}$ is strictly decreasing and strictly convex on $[0, C]$, with $D_j'(b) = -(2\ln 2)\,D_j(b)\,\varphi'(b) < 0$ and $D_j''(b) = (2\ln 2)\,D_j(b)\big[(2\ln 2)\varphi'(b)^2 - \varphi''(b)\big] > 0$. Consequently, $\lb$ is a separable, strictly convex program on $[0, C]^d$ and the budget constraint binds (for $m \le d$). Since $-D_j'(b) = (2\ln 2)\,D_j(b)\,\varphi'(b)$, absorbing $2\ln 2$ into the multiplier $\lambda > 0$ gives the equal-slope KKT conditions:
\begin{equation}
\label{eq:crwf-kkt}
\begin{cases}
b_j = 0 & \text{if }\; \sigma_j^2\,\varphi'(0) \le \lambda, \\
b_j = C & \text{if }\; D_j(C)\,\varphi'(C) \ge \lambda, \\
D_j(b_j)\,\varphi'(b_j) = \lambda & \text{otherwise.}
\end{cases}
\end{equation}
Let $\sigma_1^2 \ge \dots \ge \sigma_d^2$. 
Since $D_j(b)\varphi'(b)$ is strictly decreasing in $b$, the active coordinates ($\sigma_j^2 > \lambda/\varphi'(0)$) form a prefix $[d']$ with monotone allocations $C \ge b_1 \ge \dots \ge b_{d'} > 0$ (and $b_j = 0$ for $j > d'$). For a fixed $\lambda$, finding each $b_j \in [0, C]$ is fast via binary search or Newton's method (or in closed form if possible). An outer binary search finds $\lambda$ for which $\sum_{j=1}^{d'} b_j = mC$ (or $\sum_{j\in g} b_j = C$ when solving for $e_{\ez}^2(g)$ on a single segment $g$).

\subsection{Bounding the Bit Packing Error}
\label{sec:algorithm}
\label{sec:approx-factor}

\ezpq partitions the $d'$ active coordinates into $m$ segments. 
Each segment's bit allocation sums up to roughly $C$. Intuitively, if the packing is good, \ezpq's cost should then be close to $\lb$.
We make this precise in three steps.
First, we collect basic facts about packing the items $b^{\lb}_j$ into $m$ bins of capacity $C$ (Section~\ref{sec:packing-facts}).
Second, we quantify the change in distortion caused by over-packing (or under-packing) a single segment (Section~\ref{sec:single-segment}).
Third, we sum over segments. Because $\lb$ equalizes the marginal slopes of all coordinates and the discrepancies sum to zero, the first-order effects of over-packed and under-packed segments cancel exactly, leaving an error that is second order in the discrepancies (Section~\ref{sec:sum-segments}).
Finally, this second-order error yields practical packing guidelines.

\subsubsection{Facts about Locally Optimal Packing}
\label{sec:packing-facts}
This subsection uses only the fact that the items $b^{\lb}_j \in (0, C]$, $j \in [d']$, sum to $mC$.
A packing assigns them to segments $G = \{g_1, \dots, g_m\}$ of widths $d_g = |g|$.
The load of segment $g$ is $\sum_{j\in g} b^{\lb}_j = C + \delta_g$, where the \emph{packing discrepancy} $\delta_g$ is positive for over-packed segments and negative for under-packed ones. Since the total load is $mC$,
\begin{equation}
\label{eq:delta-sum}
\sum_{g} \delta_{g} \;=\; 0.
\end{equation}
A packing is \emph{locally optimal} if no move of a single item from one segment to another reduces the larger of the two segment loads.
A move of an item of size $b$ from $g$ to $g'$ does so if and only if $\delta_g - \delta_{g'} > b$. 
Note, this is also the condition for locally reducing $\sum_g \delta_g^2$.
Any packing can easily be made locally optimal. 
Simply apply single-item improving moves until no such moves are available. 
Since each move reduces $\sum_g \delta_g^2$ the process terminates and the maximum load is not increased.

In a locally optimal packing, over-packing (positive $\delta_g$) is always bounded by the size of the smallest item in the bin.
This is easy to see. Assume by contradiction that for some segment $\delta_g \ge \min_{j\in g} b^{\lb}_j$.
By $\sum_{g} \delta_{g} = 0$ and $\delta_g > 0$, for some other segment $\delta_{g'} < 0$. Then $\delta_g - \delta_{g'} > \delta_g \ge \min_{j\in g} b^{\lb}_j$, so moving the smallest item of $g$ to $g'$ is an improving move, which contradicts the local optimality condition. 
We conclude that
\begin{equation}
\label{eq:delta-smaller-than-min}
\delta_g < \min_{j\in g} b^{\lb}_j \ . 
\end{equation}
Consequently, every item of size $C$ occupies its own segment, every over-packed segment has $d_g \ge 2$, and $\delta_g < C/(d_g - 1)$.
From now on, we assume that the packing is locally optimal.

\subsubsection{Bounding the Error of a Single Segment}
\label{sec:single-segment}
For a segment $g$, let $(b^{\alg}_j)_{j\in g}$ be its \ez bit allocation~\eqref{eq:seg-kkt-def}, with $\sum_{j\in g} b^{\alg}_j = C$. Define the segment costs
\[
\lb(g) \;:=\; \sum_{j\in g} D_j(b^{\lb}_j),
\qquad
\alg(g) \;:=\; \sum_{j\in g} D_j(b^{\alg}_j) \;=\; d_g\, e^2(g).
\]
To bound the ratio between $\alg(g)$ and $\lb(g)$ we upper bound $\alg(g)$ using a feasible perturbation of $b^{\lb}_g$.

\begin{lemma}[Segment Bound]
\label{lem:seg-cost-bound}
Let $H_g := d_g / \sum_{j\in g}(1/\varphi'(b^{\lb}_j)) \le 1$ be the harmonic mean of $\varphi'$ in $g$. 
Set $\varepsilon_g := H_g\,\delta_g/d_g$ (and $\varepsilon_{g_0} := 0$).
Then, every segment satisfies
\begin{equation}
\label{eq:seg-cost-bound}
\alg(g) \;\le\; \lb(g)\cdot 2^{2\varepsilon_g}.
\end{equation}
This holds for both positive and negative $\delta_g$. 
\end{lemma}

\begin{proof}
The basic idea of the proof is to subtract (or add) $p_j \delta_g$ bits from $b^{\lb}_j$ to arrive at a feasible solution $\tilde{b}$ for the within-segment allocation. 
Let $p_j \ge 0$ sum up to $1$ and set $\tilde{b}_j := b^{\lb}_j - p_j\delta_g$.
Then $\sum_{j\in g}\tilde{b}_j = C$ because $\sum_{j\in g} p_j = 1$.
Moreover, if $\delta_g > 0$, then $p_j \le 1$ and from equation~\eqref{eq:delta-smaller-than-min} we get that $\min_{j \in g} \tilde{b}_j \ge \min_{j \in g} b^{\lb}_j - \delta_g > 0$; if $\delta_g \le 0$, then $\tilde{b}_j \ge b^{\lb}_j > 0$ directly. In both cases $\tilde{b}_j > 0$ for all $j \in g$, and together with $\sum_{j\in g}\tilde{b}_j = C$ this gives $\tilde{b}_j \le C$, so $\tilde{b}$ lies in the domain $[0, C]$ of $\varphi$.
Hence $\tilde{b}$ is feasible for~\eqref{eq:seg-kkt-def} and $\alg(g) \le \sum_{j\in g} D_j(\tilde{b}_j)$.

By convexity of $\varphi$, for either sign of $\delta_g$, we get 
$\varphi(\tilde{b}_j) \ge \varphi(b^{\lb}_j) - \varphi'(b^{\lb}_j)\,p_j\delta_g$. 
Setting $p_j := (1/\varphi'(b^{\lb}_j))/\sum_{i\in g}(1/\varphi'(b^{\lb}_i))$ to be inversely proportional to $\varphi'(b^{\lb}_j)$ gives $\varphi(\tilde{b}_j) \ge \varphi(b^{\lb}_j) - H_g\delta_g/d_g$. 
\[
\sum_{j \in g} D_j(\tilde{b}_j) = \sum_{j \in g} D_j(b^{\lb}_j)\,2^{2(\varphi(b^{\lb}_j) - \varphi(\tilde{b}_j))} \le \sum_{j \in g} D_j(b^{\lb}_j)\,2^{2\varepsilon_g} = \lb(g)\, 2^{2\varepsilon_g}\ .
\]
Combining the above gives  $\alg(g) \le \lb(g)\, 2^{2\varepsilon_g}$.
\end{proof}

The value of $\varepsilon_g$ could potentially be large.
Consider $m+1$ equal-variance coordinates, to which $\lb$ assigns $mC/(m+1)$ bits each.
Every partition must place two coordinates in one segment. Hence the locally optimal packing (one pair and $m-1$ singletons) attains $\mathrm{OPT}$ exactly. This gives $\varepsilon_g \approx C/2$ as $m$ grows. The exponent $2\varepsilon_g$ in~\eqref{eq:seg-cost-bound} can therefore be close to $C$ even for a segment of an optimal packing. 
If, on the other hand, $\max_j b^{\lb}_j \le C/t$ for some integer $t \ge 1$, then $\alg/\lb \le 2^{\frac{2C}{t(t+1)}}$.
In the next section we show that a much better bound is possible. 

\subsubsection{Summing over Segments: First-Order Cancellation}
\label{sec:sum-segments}

The last section ignores the fact that $\alg/\lb$, written as~\eqref{eq:weighted-avg} below, is an average, in which over-packed segments ($\varepsilon_g > 0$) are offset by under-packed ones ($\varepsilon_g < 0$).
The following lemma shows that this offset is exact to first order: the weighted average of the exponents vanishes.
Define the segment weights $w_g := \lb(g)/\lb$, so that $\sum_g w_g = 1$. By Lemma~\ref{lem:seg-cost-bound},
\begin{equation}
\label{eq:weighted-avg}
\frac{\alg}{\lb} \;=\; \sum_{g} w_g\, \frac{\alg(g)}{\lb(g)} \;\le\; \sum_{g} w_g\, 2^{2\varepsilon_g}.
\end{equation}

\begin{lemma}[First-Order Cancellation]
\label{lem:first-order}
$\sum_g w_g\,\varepsilon_g = 0$. Moreover, $w_g\,\varepsilon_g^2 = \frac{\lambda}{\lb}\cdot\frac{H_g\,\delta_g^2}{d_g} \le \frac{\delta_g^2}{d'\,d_g}$ for every segment.
\end{lemma}

\begin{proof}
By the KKT conditions~\eqref{eq:crwf-kkt} and local optimality (Section~\ref{sec:packing-facts}), every $j \in g$ is either interior ($0 < b^{\lb}_j < C$) or a capped singleton. For interior coordinates, the KKT conditions~\eqref{eq:crwf-kkt} give $D_j(b^{\lb}_j) = \lambda/\varphi'(b^{\lb}_j)$, so $\lb(g) = \lambda\sum_{j\in g} 1/\varphi'(b^{\lb}_j) = \lambda\, d_g/H_g$ and therefore $w_g\,\varepsilon_g = \lambda\,\delta_g/\lb$. This identity also holds for capped singletons and for $g_0$, where $\varepsilon_g = \delta_g = 0$. By~\eqref{eq:delta-sum}, $\sum_g w_g\,\varepsilon_g = \frac{\lambda}{\lb}\sum_{r=1}^m \delta_{g_r} = 0$.
Multiplying the identity by $\varepsilon_g = H_g\delta_g/d_g$ gives $w_g\,\varepsilon_g^2 = \frac{\lambda}{\lb}\cdot\frac{H_g\,\delta_g^2}{d_g}$. Finally, every active coordinate has $D_j(b^{\lb}_j) \ge \lambda/\varphi'(b^{\lb}_j) \ge \lambda$ (with equality in the first step for interior coordinates, and by~\eqref{eq:crwf-kkt} for capped ones), so $\lb \ge \lambda d'$. Together with $H_g \le 1$, this gives $w_g\,\varepsilon_g^2 \le \delta_g^2/(d'\,d_g)$.
\end{proof}

\begin{theorem}[Second-Order Packing Bound]
\label{thm:second-order}
For any locally optimal packing with $C \le 8$,
\begin{equation}
\label{eq:second-order}
\frac{\alg}{\lb} \;\le\; 1 + \frac{16}{d'}\sum_{g} \frac{\delta_{g}^2}{d_{g}}.
\end{equation}
\end{theorem}

\begin{proof}
First, $\varepsilon_g < C/2$ for every segment. This is trivial if $\delta_g \le 0$. If $\delta_g > 0$, then $d_g \ge 2$ and $\delta_g < C/(d_g - 1)$ by local optimality (Section~\ref{sec:packing-facts}), so $H_g \le 1$ gives $\varepsilon_g \le \delta_g/d_g < C/(d_g(d_g-1)) \le C/2$.
Next, $2^x \le 1 + x\ln 2 + 4x^2$ for all $x \le 8$.
Applying this inequality to~\eqref{eq:weighted-avg} with $x = 2\varepsilon_g < C \le 8$ and using Lemma~\ref{lem:first-order},
\[
\frac{\alg}{\lb} \;\le\; \sum_g w_g\,\big(1 + 2\ln 2\cdot\varepsilon_g + 16\varepsilon_g^2\big) \;=\; 1 + 16\sum_g w_g\,\varepsilon_g^2 \;\le\; 1 + \frac{16}{d'}\sum_{g} \frac{\delta_{g}^2}{d_{g}}. \qedhere
\]
\end{proof}
The constant $16$ accounts for the worst case $\varepsilon_g \to C/2$; for small $\varepsilon_g$ it can be replaced by $2(\ln 2)^2 \approx 0.96$.
For $C > 8$ it should be replaced by a larger constant. 

\paragraph{Practical Packing Guidelines}
The analysis above also gives three \emph{guidelines} for the packer:
(i) \emph{minimize $\sum_g \delta_g^2/d_g$} which minimizes the second-order term of Theorem~\ref{thm:second-order};
(ii) positive spillage should go to wide segments (large $d_g$), where $\delta_g^2/d_g$ is small, rather than to narrow head segments (also, $H_g$ is smallest at low bit rates); and
(iii) \emph{finish with local search}: applying single-item moves that reduce $\sum_g \delta_g^2$ until none remains makes the packing locally optimal, which is the only assumption of Lemma~\ref{lem:seg-cost-bound} and Theorem~\ref{thm:second-order}.

\paragraph{Instance-wise Certification}
Rather than relying on these bounds, \ezpq computes $\lb$ directly and outputs the value of $\alg/\lb$ as a certificate. Both $\alg$ and $\lb$ are computed analytically from the spectrum under the \ez model, before any $k$-means is run. The certificate therefore bounds the loss caused by the segmentation (the bit packing) within the Gaussian \ez model. It is not a bound on the exact Gaussian or on the empirical $k$-means distortion, which can deviate from the \ez estimate by a few percent (Appendix~\ref{app:coupled-zador-validation}).

\section{The \ezpq Algorithm}
Algorithm~\ref{alg:sopq} summarizes \ezpq. 
It does not provide details about creating the spline $\varphi$ with \code{effective\_bit\_rate}, computing the global bit allocation by LB \code{global\_bit\_allocation}, the packing algorithm \code{bin\_pack\_segments}, or $k$ means clustering \code{k\_means\_centers}.
These are described above in sufficient detail for practitioners to apply them correctly.
Note that the algorithm uses the empirical covariance of the input data. Its encode and decode functions are identical to PQ.

It is worth noting an implementation detail that was not discussed above.
It is convenient to have the function \code{bin\_pack\_segments} return a permutation \code{pi} of $[d]$ together with the segment range slices $\code{seg}$. Applying the permutation (and truncation) to the preconditioner \code{U' <- U[:, pi[1:d']]} serves two purposes. 
First, \code{pi} need not be stored in the model.
But more importantly, for rotated vectors, \code{y[seg[r]]} are contiguous slices in memory, eliminating index gathers and making $k$-means, encoding and query-time LUT inner products SIMD-friendly.
Truncating zero-bit coordinates (those in $g_0$) also reduces rotation memory and projection cost from $O(d^2)$ to $O(d \cdot d')$. 

\begin{lstlisting}[style=sopqalg,escapeinside={(*@}{@*)},caption={The \ezpq algorithm.},label={alg:sopq}]
fit(X, k, m):                                                                                   
  C = log2(k); phi = effective_bit_rate(C)                         # per segment bit budget C, effective bit rate function phi
  mu <- mean(X);  U, s2 <- eig(covariance(X - mu))                 # Eigenvectors and eigenvalues of the empirical covariance
  b, d' <- global_bit_allocation(s2, m, C, phi)                    # b are the global bit allocations by LB, d' active coordinates
  pi, seg <- bin_pack_segments(b, m)                               # pi is a segment-contiguous permutation, seg are slices
  alg, lb = compute_cert(b, s2, pi, seg, phi, C); cert <- alg/lb   # cert is the ratio between EZ alg cost and the lower bound
  U' <- U[:, pi[1:d']]                                             # Permute U by pi and trim inactive coordinates. U' is d*d'
  Y <- U'^T (X - mu)                                               # Y is segment-contiguous and d' dimensional 
  cs <- [k_means_centers(Y[seg[r], :], k) for r=1..m]              # cs contains k centers for each of the m segments 
  return model = (mu, U', seg, cs, cert)

encode(x):                                                         # Identical to PQ
  y <- U'^T (x - mu)                                               # Remove mean, project onto eigenbasis (implicitly permute)
  return code = [argmin j: norm(y[seg[r]] - cs[r][j]) for r=1..m]  # Return m closest center indexes. One per segment

reconstruct(code):                                                 # Identical to PQ
  for r=1..m:  y_hat[seg[r]] <- cs[r][code[r]]                     # y_hat reconstructed from center vectors in cs
  return x_hat = mu + U' y_hat                                     # x_hat is unrotated and the mean added back
\end{lstlisting}

\section{Experimental Evaluation}
\label{sec:experiments}

We evaluate \ezpq in three experiments:

\begin{enumerate}
  \item \textbf{Packing comparison (Section~\ref{sec:results-packing}).} We compare the \ez cost of twelve packers of the \lb bit allocations on seven synthetic Gaussian spectra.
  \item \textbf{Certifying optimality on real embeddings (Section~\ref{sec:results-certificates}).} We compare the \ez cost of \ezpq against five baselines on PCA spectra of six embedding datasets from VIBE~\citep{jaasaari2025}.
  \item \textbf{End-to-end MSE and recall on VQ-bench (Section~\ref{sec:vq-bench-results}).} We evaluate \ezpq against 15 baseline quantizers including OPQ-P~\citep{ge2013opq}, PQ~\citep{jegou2011pq}, EDEN~\citep{vargaftik2022eden}, E-RaBitQ~\citep{gao2024erabitq}, and TurboQuant~\citep{zandieh2025turboquant} in VQ-bench~\citep{padaki2026vqbench}.
\end{enumerate}

\subsection{Packing under the \ez Model}
\label{sec:results-packing}

We evaluate the 12 packer variants specified in Table~\ref{tab:packers} by the exact \ez cost of their packings of the \lb bit allocations~\eqref{eq:crwf-kkt} on $7$ synthetic Gaussian spectra (three power laws of different steepness, an exponential decay, a head spike with a slow tail, a step spectrum, and a high-dimensional power law). Throughout, we use the spline $\varphi$ of Appendix~\ref{app:zador-approximations} and $C=8$. We use the rates $\bar{b} \in \{0.5, 1, 2, 4\}$ bits/dim for all spectra except for the high-dimensional power law, where we use $\bar{b} \in \{1/8, 1/4, 1/2\}$. Table~\ref{tab:suite1-packers} reports the approximation quality $\mathrm{ALG}/\lb - 1$ of each evaluated method, along with the weighted absolute packing exponent $\E_W[|\varepsilon_g|]$.

\begin{table}[!htbp]
\caption{Packers compared in Section~\ref{sec:results-packing}. Each packer partitions the active coordinates of the \lb bit allocation~\eqref{eq:crwf-kkt} into $m$ bins. Every coordinate with $b_j = C$ first takes a width-$1$ bin of its own, and the packer partitions the rest into the remaining bins (with the exception of \code{lpt\_fixed}). Following any algorithm with cost based local search \code{*\_ls} makes it locally optimal with respect to $\sum_g w_g 2^{2\varepsilon_g}$. Note that this is distinct from the local optimality of Section~\ref{sec:packing-facts} with respect to $\delta_g$ which some packing algorithm (like \code{lpt}) already exhibit.} 
\label{tab:packers}
\centering
\footnotesize
\setlength{\tabcolsep}{4pt}
\begin{tabular}{@{}lp{0.78\linewidth}@{}}
\toprule
\textbf{Packer} & \textbf{Rule} \\
\midrule
\code{lpt\_fixed} & OPQ-style allocation (ambient OPQ cap $d_g \le \lceil d/m \rceil$): place each item in order $b_1 \ge \dots \ge b_{d'}$ into the least full unsealed bin ($|g| < \lceil d/m \rceil$), sealing bins at $\lceil d/m \rceil$ items. \\
\code{lpt} & Longest processing time~\citep{graham1969,chandra1975}: place item large-to-small into the least full bin with no size cap. \\
\code{online\_greedy} & Greedy algorithm: seed the $m$ bins with $b_1, \dots, b_m$, then place each subsequent item $j$ into the bin yielding the smallest marginal increase in $w_g 2^{2\varepsilon_g}$, under the online bit target $C_t = (\sum_{j \le t} b_j)/m$. \\
\code{in\_order} & Contiguous threshold cut: sweep $b_1 \ge \dots \ge b_{d'}$, taking at least one item per bin and adding to the current contiguous bin while $w_g 2^{2\varepsilon_g}$ $\le \tau$. Binary search for the lowest possible $\tau$. \\
\code{prefix\_cut} & Equal-bit contiguous cut: sweep $b_1 \ge \dots \ge b_{d'}$ and place the $r$-th cut ($r = 1, \dots, m-1$) at the index with cumulative bit sum closest to $rC$ (ensuring $\ge 1$ item per bin). \\
\code{contiguous\_dp} & Contiguous DP: dynamic programming over $[d']$ to find the minimum $\sum_g w_g 2^{2\varepsilon_g}$ contiguous partition into $m$ bins in $O(m d'^2)$ time. \\
\code{*\_ls} & Cost based local search appended to any base packer: iteratively apply $1$-item moves and $1$-for-$1$ swaps across bins whenever doing so strictly decreases $\sum_g w_g 2^{2\varepsilon_g}$. \\
\bottomrule
\end{tabular}
\end{table}

\begin{table}[!htbp]
\caption{Packing \lb bit allocations under the \ez model: excess $\mathrm{ALG}/\lb - 1$ (\%) over 27 synthetic instances (21 at $\bar{b} \le 2$, 6 at $\bar{b} = 4$) with $C = 8$. $\E_w[|\varepsilon_g|] = \sum_g w_g |\varepsilon_g|$ is the absolute exponent $|\varepsilon_g|$ of Lemma~\ref{lem:seg-cost-bound} (bits) weighted by $w_g$ and averaged over instances. Bold: best per column.}
\label{tab:suite1-packers}
\centering
\footnotesize
\setlength{\tabcolsep}{4.5pt}
\begin{tabular}{@{}lrrrrr@{}}
\toprule
 & \multicolumn{2}{c}{$\bar{b} \le 2$ b/dim} & \multicolumn{2}{c}{$\bar{b} = 4$ b/dim} & \\
\cmidrule(lr){2-3}\cmidrule(lr){4-5}
\textbf{Packer} & \textbf{Mean} & \textbf{Max} & \textbf{Mean} & \textbf{Max} & $\E_w[|\varepsilon_g|]$ \\
\midrule
\code{lpt\_fixed} & 0.51 & 4.34 & 17.0 & 32.6 & 0.0815 \\
\code{lpt} & 0.28 & 2.52 & 9.62 & 19.6 & 0.0647 \\
\code{online\_greedy} & 0.30 & 2.52 & 9.14 & \textbf{17.5} & 0.0650 \\
\code{in\_order} & 7.55 & 29.5 & 43.6 & 75.8 & 0.2968 \\
\code{prefix\_cut} & 7.84 & 33.5 & 113 & 283 & 0.2357 \\
\code{contiguous\_dp} & 3.94 & 24.9 & 41.0 & 69.4 & 0.1961 \\
\midrule
\code{lpt\_fixed\_ls} & 0.31 & 4.23 & 16.9 & 32.3 & 0.0682 \\
\code{lpt\_ls} & \textbf{0.05} & 0.94 & 7.28 & \textbf{17.5} & 0.0441 \\
\code{online\_greedy\_ls} & 0.06 & 0.93 & 7.21 & \textbf{17.5} & 0.0439 \\
\code{in\_order\_ls} & 0.07 & \textbf{0.72} & \textbf{6.91} & \textbf{17.5} & 0.0459 \\
\code{prefix\_cut\_ls} & 0.08 & \textbf{0.72} & 6.98 & \textbf{17.5} & 0.0466 \\
\code{contiguous\_dp\_ls} & 0.09 & 0.81 & 6.98 & \textbf{17.5} & 0.0474 \\
\bottomrule
\end{tabular}
\end{table}
\FloatBarrier

Without local search, \code{lpt} performs best for $\bar{b} \le 2$ ($0.28\%$ mean excess), closely followed by \code{online\_greedy}. Local search yields the best performance overall, with \code{lpt\_ls} obtaining $0.05\%$ overhead on average, and \code{in\_order\_ls} and \code{prefix\_cut\_ls} $0.72\%$ overhead at most, for $\bar{b} \le 2$. At $\bar{b}=4$, five out of the six local-search variants reach the same $17.5\%$ maximum, which is the exact optimum on that instance.

\subsection{Certified Optimality on Real Embedding Spectra}
\label{sec:results-certificates}

We evaluate \ezpq against five baselines (Table~\ref{tab:suite2-methods}) on the PCA spectra of six embedding datasets from VIBE~\citep{jaasaari2025} (Table~\ref{tab:suite2-datasets}). 
Each spectrum is computed from a different sample of $25{,}000$ base vectors. As in Section~\ref{sec:results-packing}, we evaluate each method by its \ez cost and report the excess $\mathrm{ALG}/\lb - 1$ over the lower bound $\lb$ of each spectrum.
For PQ and random rotation, whose segments are not axis-aligned in the PCA basis, the \ez cost of a segment is computed from the eigenvalues of the corresponding principal block of the covariance matrix. All certificates in this section are relative to the Gaussian \ez model. 
They measure the loss caused by the segmentation, and account neither for the estimation error of \ez (Appendix~\ref{app:coupled-zador-validation}) nor for the non-Gaussianity of real embeddings.

\begin{table}[!htbp]
\caption{Product quantization baselines compared in this section.}
\label{tab:suite2-methods}
\centering
\footnotesize
\setlength{\tabcolsep}{4pt}
\begin{tabular}{@{}llp{0.62\linewidth}@{}}
\toprule
\textbf{Method} & \textbf{Rotation} & \textbf{Segments} \\
\midrule
\ezpq & PCA & \lb bits packed by \code{lpt}; zero-bit coordinates pruned. \\
OPQ-P & PCA & Equal widths $d/m$; eigenvalues allocated to balance the product of variances per segment~\citep{ge2014opqJournal}. \\
PCA-PQ & PCA & Equal widths $d/m$, contiguous in eigenvalue order. \\
PQ & None & Equal widths $d/m$, contiguous in the original coordinates~\citep{jegou2011pq}. \\
Random rotation & Haar-random & Equal widths $d/m$, contiguous. \\
Uniform diagonal & - & Equal widths $d/m$ at $s^2 = \bar{\sigma}^2\mathbf{1}$ (expected spectrum of random rotation). \\
\bottomrule
\end{tabular}
\end{table}

\begin{table}[!htbp]
\caption{Embedding datasets from VIBE~\citep{jaasaari2025}.}
\label{tab:suite2-datasets}
\centering
\footnotesize
\setlength{\tabcolsep}{5pt}
\begin{tabular}{@{}llrr@{}}
\toprule
\textbf{Dataset} & \textbf{VIBE name} & $d$ & \textbf{Base vectors} \\
\midrule
Llama & \code{llama-128-ip} & 128 & 256{,}921 \\
Yahoo & \code{yahoo-minilm-384-normalized} & 384 & 677{,}305 \\
ImageNet & \code{imagenet-clip-512-normalized} & 512 & 1{,}281{,}167 \\
COCO & \code{coco-nomic-768-normalized} & 768 & 282{,}360 \\
CCNews & \code{ccnews-nomic-768-normalized} & 768 & 495{,}328 \\
ArXiv & \code{arxiv-nomic-768-normalized} & 768 & 1{,}344{,}643 \\
\bottomrule
\end{tabular}
\end{table}

Table~\ref{tab:suite2-certificates} compares \ezpq (\lb packed via \code{lpt}) with OPQ-P, both measured against the same $\lb$.
Table~\ref{tab:suite2-results} reports the mean excess ($\mathrm{ALG}/\lb-1$) of each method at each rate.
For $\bar{b} \le 2$, both \ezpq and OPQ-P are nearly optimal under \ez: both are within $0.66\%$ of $\lb$ on every dataset, and for $\bar{b} \le 1$ within $0.13\%$. \ezpq has lower or equal excess on all six datasets for $\bar{b} \le 1$ and on five of them for $\bar{b} = 2$, but by at most $0.11$ percentage points. For $\bar{b} = 4$, both are $0.5$-$2.9\%$ above $\lb$ and within $0.2$ percentage points of each other. There is a clear separation between the nontrivial allocation methods (\ezpq and OPQ-P) and the baselines (PCA-PQ, PQ, random rotation, uniform diagonal); for example, \ezpq and OPQ-P achieve $0.04\%$ and $0.06\%$ mean excess across datasets for $\bar{b}=1$, whereas the other baselines suffer $185$-$213\%$ at the same bit-rate. The remaining excess of \ezpq for $\bar{b} \le 2$ is a packing loss: adding the local search of Section~\ref{sec:results-packing} to its \code{lpt} packing (\ezpq{} + \code{lpt\_ls} in Table~\ref{tab:suite2-results}) reduces it below $0.001\%$ on every dataset.

\begin{table}[!htbp]
  \caption{$\mathrm{ALG}/\lb - 1$ (\%) on real embedding spectra for \ezpq (\lb bits packed by \code{lpt}) and Parametric OPQ (OPQ-P), under the \ez model. Best per dataset and rate in bold.}
  \label{tab:suite2-certificates}
\centering
\footnotesize
\setlength{\tabcolsep}{4.5pt}
\begin{tabular}{@{}lrrrrrrrrr@{}}
\toprule
 & & \multicolumn{2}{c}{$\bar{b} = 0.5$} & \multicolumn{2}{c}{$\bar{b} = 1$} & \multicolumn{2}{c}{$\bar{b} = 2$} & \multicolumn{2}{c}{$\bar{b} = 4$} \\
\cmidrule(lr){3-4}\cmidrule(lr){5-6}\cmidrule(lr){7-8}\cmidrule(lr){9-10}
\textbf{Dataset} & $d$ & \ezpq & OPQ-P & \ezpq & OPQ-P & \ezpq & OPQ-P & \ezpq & OPQ-P \\
\midrule
Llama & 128 & \textbf{0.00} & 0.01 & \textbf{0.03} & 0.07 & \textbf{0.45} & 0.50 & \textbf{0.47} & \textbf{0.47} \\
Yahoo & 384 & \textbf{0.00} & \textbf{0.00} & \textbf{0.03} & \textbf{0.03} & 0.66 & \textbf{0.65} & \textbf{2.79} & \textbf{2.79} \\
ImageNet & 512 & \textbf{0.00} & 0.01 & \textbf{0.01} & 0.05 & \textbf{0.16} & 0.20 & 0.87 & \textbf{0.77} \\
COCO & 768 & \textbf{0.00} & 0.01 & \textbf{0.00} & 0.03 & \textbf{0.21} & 0.29 & \textbf{2.77} & 2.94 \\
CCNews & 768 & \textbf{0.00} & 0.01 & \textbf{0.07} & \textbf{0.07} & \textbf{0.20} & 0.22 & \textbf{1.08} & \textbf{1.08} \\
ArXiv & 768 & \textbf{0.00} & 0.04 & \textbf{0.06} & 0.13 & \textbf{0.24} & 0.35 & \textbf{1.69} & 1.79 \\
\bottomrule
\end{tabular}
\end{table}

\begin{table}[!htbp]
\caption{Mean $\mathrm{ALG}/\lb - 1$ (\%) of each method over the six datasets, under the \ez model with $C = 8$. \ezpq{} + \code{lpt\_ls} adds the local search of Section~\ref{sec:results-packing} to the \code{lpt} packing of \ezpq.}
\label{tab:suite2-results}
\centering
\footnotesize
\setlength{\tabcolsep}{5pt}
\begin{tabular}{@{}lrrrr@{}}
\toprule
 & \multicolumn{4}{c}{\textbf{Mean excess over $\mathrm{LB}$ (\%)}} \\
\cmidrule(lr){2-5}
\textbf{Method} & $\bar{b} = 0.5$ & $\bar{b} = 1$ & $\bar{b} = 2$ & $\bar{b} = 4$ \\
\midrule
\ezpq & 0.00 & 0.04 & 0.32 & 1.61 \\
\ezpq{} + \code{lpt\_ls} & 0.00 & 0.00 & 0.00 & 1.48 \\
OPQ-P & 0.01 & 0.06 & 0.37 & 1.64 \\
PCA-PQ & 138 & 202 & 251 & 255 \\
PQ & 122 & 185 & 235 & 243 \\
Random rotation & 129 & 195 & 246 & 252 \\
Uniform diagonal & 157 & 213 & 257 & 257 \\
\bottomrule
\end{tabular}
\end{table}

\subsection{End-to-end VQ-bench Results}
\label{sec:vq-bench-results}

We benchmark \ezpq against the 15 baseline quantizers of \vqb~\citep{padaki2026vqbench} on ten VIBE datasets~\citep{jaasaari2025} ($d$ from $128$ to $1024$). These are the six datasets of Table~\ref{tab:suite2-datasets} together with HotpotQA, LAION, Landmark, and MS MARCO. Every method is fit on $25{,}000$ database vectors. Recall@$10$ is measured on $1{,}000$ evaluation queries, each against its $100$ candidates and the reconstruction MSE (the average squared distance between a vector and its reconstruction) on $1{,}000$ sampled database vectors. \ezpq uses the \lb bits of the spline $\varphi$, packed by \code{lpt} as in Section~\ref{sec:results-certificates}. Unlike Section~\ref{sec:ezg-validation}, which trains codebooks with $k$-means++ and a Lloyd-Max quantizer for one-dimensional segments, \vqb trains the $k = 256$-centroid codebooks of \ezpq, PQ, OPQ and OPQ-P with its own $k$-means (random initialization, $25$ Lloyd iterations). All results are from a single run with one seed. We did not measure the seed-to-seed variance. Differences of fractions of a percent between \ezpq, OPQ-P and OPQ should not be read as significant.

Table~\ref{tab:vqbench-recon} reports reconstruction MSE at $0.5$, $1$, $2$ and $4$ code bits per dimension, and Table~\ref{tab:vqbench-recall} in Appendix~\ref{app:vqbench-recall} reports recall@$10$. PQ, OPQ, OPQ-P and \ezpq were run at exactly these code rates. Every other quantizer is shown at its configuration of the same nominal bit width, which uses up to $0.75$ more code bits per dimension to store per-vector norms. This favors those baselines. \ezpq has a lower MSE than those baselines on every dataset and rate despite the handicap. Appendix~\ref{app:vqbench-timing} reports encode and scoring times.

We note that, for clarity of presentation, the reported bit rates for PQ-family methods exclude model overhead (e.g., the mean, rotation matrix, and PQ codebooks). This overhead is negligible at scale, as the model is stored only once and shared across all vectors. Specifically, when stored in float32, the model contributes at most $32(d+k+1)/n$ additional bits per dimension, contributing only about $0.03$ bits per dimension for $d=768$ and $n=10^6$.

\begin{table}[!htbp]
\caption{Reconstruction MSE on the ten \vqb datasets at $\bar{b}$ code bits per dimension (excluding model overhead); $\times 10^{3}$ for the nine datasets of unit-norm vectors (all but Llama). \emph{Best other} is the lowest MSE among the $12$ quantizers outside the PQ family at the same nominal bit width: SimHash at $\bar{b} = 0.5$, EDEN-MSE or ITQ at $\bar{b} = 1$, and EDEN-MSE at $\bar{b} \in \{2, 4\}$. Bold: best per dataset and rate.}
\label{tab:vqbench-recon}
\centering
\footnotesize
\setlength{\tabcolsep}{3.5pt}
\resizebox{\linewidth}{!}{%
\begin{tabular}{@{}llrrrrrrrrrr@{}}
\toprule
$\bar{b}$ & \textbf{Method} & \textbf{Llama} & \textbf{Yahoo} & \textbf{ImageNet} & \textbf{COCO} & \textbf{CCNews} & \textbf{ArXiv} & \textbf{HotpotQA} & \textbf{LAION} & \textbf{Landmark} & \textbf{MS MARCO} \\
 & & $d = 128$ & $d = 384$ & $d = 512$ & $d = 768$ & $d = 768$ & $d = 768$ & $d = 640$ & $d = 512$ & $d = 768$ & $d = 1024$ \\
\midrule
$0.5$ & \ezpq & 29.2 & \textbf{359} & 137 & 37.9 & \textbf{99.2} & \textbf{88.4} & \textbf{258} & 197 & 38.8 & \textbf{211} \\
 & OPQ-P & 29.2 & 360 & 138 & 38.0 & 99.5 & 88.8 & \textbf{258} & 198 & 39.1 & 212 \\
 & OPQ & \textbf{28.8} & 360 & \textbf{135} & \textbf{37.2} & 100 & 89.4 & 259 & \textbf{194} & \textbf{38.5} & 215 \\
 & PQ & 43.2 & 527 & 214 & 92.3 & 238 & 219 & 357 & 261 & 99.6 & 451 \\
 & Best other & 82.4 & 696 & 364 & 154 & 322 & 313 & 477 & 416 & 166 & 586 \\
\midrule
$1$ & \ezpq & 14.5 & \textbf{181} & 71.8 & \textbf{18.2} & \textbf{42.5} & \textbf{37.9} & \textbf{131} & \textbf{110} & \textbf{19.0} & \textbf{98.1} \\
 & OPQ-P & 14.5 & \textbf{181} & 71.8 & 18.3 & 42.6 & 38.1 & 132 & \textbf{110} & 19.1 & 98.5 \\
 & OPQ & \textbf{14.4} & 182 & \textbf{71.0} & 18.4 & 42.8 & 38.4 & 133 & 111 & 19.2 & 100 \\
 & PQ & 27.4 & 307 & 131 & 60.8 & 141 & 133 & 206 & 156 & 65.1 & 259 \\
 & Best other & 40.1 & 359 & 178 & 68.9 & 166 & 161 & 246 & 214 & 75.8 & 303 \\
\midrule
$2$ & \ezpq & \textbf{4.16} & \textbf{52.1} & 22.4 & \textbf{5.57} & \textbf{10.8} & \textbf{9.46} & 38.4 & 33.9 & \textbf{5.87} & \textbf{27.4} \\
 & OPQ-P & \textbf{4.16} & \textbf{52.1} & \textbf{22.3} & \textbf{5.57} & \textbf{10.8} & 9.50 & \textbf{38.3} & \textbf{33.8} & \textbf{5.87} & 27.5 \\
 & OPQ & 4.17 & 52.4 & \textbf{22.3} & 5.65 & 11.0 & 9.64 & 38.9 & 34.2 & 5.96 & 28.1 \\
 & PQ & 9.31 & 94.9 & 42.6 & 20.6 & 44.3 & 42.3 & 64.2 & 49.9 & 22.0 & 79.9 \\
 & Best other & 13.6 & 116 & 60.7 & 25.6 & 53.4 & 52.1 & 79.2 & 69.1 & 27.8 & 97.9 \\
\midrule
$4$ & \ezpq & 0.415 & \textbf{5.15} & 2.51 & 0.655 & \textbf{1.14} & 0.985 & \textbf{3.82} & 3.47 & \textbf{0.685} & \textbf{2.75} \\
 & OPQ-P & 0.415 & \textbf{5.15} & \textbf{2.50} & \textbf{0.650} & \textbf{1.14} & \textbf{0.983} & \textbf{3.82} & \textbf{3.46} & 0.688 & \textbf{2.75} \\
 & OPQ & \textbf{0.414} & 5.27 & 2.59 & 0.677 & 1.18 & 1.02 & 3.93 & 3.58 & 0.716 & 2.88 \\
 & PQ & 0.959 & 8.89 & 4.28 & 2.00 & 4.13 & 3.94 & 6.04 & 5.02 & 2.12 & 7.43 \\
 & Best other & 1.08 & 9.23 & 4.88 & 2.06 & 4.29 & 4.21 & 6.39 & 5.57 & 2.24 & 7.90 \\
\bottomrule
\end{tabular}}
\end{table}
\FloatBarrier

\ezpq lowers the MSE of PQ by $25$-$78\%$ on every dataset at every rate, and its MSE is lower than that of every quantizer outside the PQ family on all ten datasets at every rate. \ezpq and OPQ-P are essentially tied: \ezpq's MSE is between $0.6\%$ below and $0.8\%$ above OPQ-P's on every dataset and rate. OPQ, which learns its rotation by alternating optimization, has a lower MSE than \ezpq on five datasets at $\bar{b} = 0.5$ and on two at $\bar{b} = 1$, by at most $1.8\%$; at $\bar{b} \ge 2$, \ezpq's MSE is lower on all but one dataset at each rate, by up to $4.3\%$. This does not contradict the certificates of Section~\ref{sec:results-certificates}, which hold under the Gaussian \ez model: real embeddings are not Gaussian, and a rotation other than the SVD basis can exploit this. \ezpq has the smallest model of the three, because it stores only $d'$ columns of the rotation: at $\bar{b} = 0.5$ its model is $39$-$62\%$ the size of OPQ-P's and $23$-$38\%$ of OPQ's. The comparison with OPQ is partly an artifact of the implementation: \vqb stores and applies OPQ's PCA initialization and its learned rotation as two separate $d \times d$ matrices. Fused offline into one, they would give OPQ the model size and scoring cost of OPQ-P. \ezpq's encode time is within $30\%$ of PQ's on every dataset and rate, and its scoring time lies between PQ's and OPQ's in $39$ of the $40$ dataset-rate pairs (Appendix~\ref{app:vqbench-timing}). Compared with OPQ-P, which like \ezpq applies a single rotation, \ezpq scores faster on all $20$ dataset-rate pairs with $\bar{b} \le 1$, where $d' < d$, and on $13$ of the $20$ pairs with $\bar{b} \ge 2$. We fix $k$ per segment because it controls the encoding and scoring cost: codes have a fixed byte-aligned layout, and scoring takes exactly $m$ table lookups. PCA-based variable-rate quantizers such as transform coding~\citep{brandt2010transform} or VAQ~\citep{paparrizos2022vaq} give up this structure. We did not benchmark them; such a comparison would separate the gain of \ezpq's multi-dimensional fixed-$k$ segments from that of PCA decorrelation and variable bit allocation alone.

\FloatBarrier

\section{Summary and Conclusion}
\label{sec:discussion}

This paper places Product Quantization (PQ) on a firmer mathematical and algorithmic footing.
First, we proved that the SVD basis $R = U^\top$ is the globally optimal preconditioning rotation for exact Gaussian $k$-means PQ (Theorem~\ref{thm:pca}).
Second, we introduced the Extended Zador (\ez) model (Section~\ref{sec:ezg}). 
It is an important standalone result providing the first vector quantization distortion model for the low bit, finite dimension, non-isotropic regime.
Third, reducing segmentation to 1-d bin packing yields \ezpq, together with an instance-specific lower bound $\lb$ under the \ez model (Lemma~\ref{thm:lb}) and an explicit second-order approximation guarantee (Theorem~\ref{thm:second-order}).
Empirically (Section~\ref{sec:experiments}), \ezpq is within $0.7\%$ of $\lb$ on real embedding spectra at up to $2$ bits per coordinate ($3\%$ at $4$ bits), matches OPQ-P's accuracy with a simpler and smaller model, and outperforms PQ ($25$--$78\%$ lower MSE) as well as all non-PQ baselines in reconstruction MSE on \vqb.

More broadly, vector quantization is becoming central to billion-scale vector search and LLM and KV-cache compression.
Strengthening the mathematical and algorithmic foundations of VQ not only explains when and why existing methods like one-shot Parametric OPQ are near-optimal, but also turns codebook and subspace design into a principled optimization problem with provable guarantees and per-instance certificates under explicit distortion models.

\FloatBarrier
\bibliography{references}
\bibliographystyle{plainnat}

\appendix
\FloatBarrier
\section{Effective Bit Rate \texorpdfstring{$\varphi(b)$}{phi(b)}}
\label{app:zador-approximations}

This appendix constructs the effective bit rate $\varphi$ used by \ez.
Before we begin, it is important to note that the algorithm is very tolerant of small discrepancies in $\varphi$.
For example, one can simply set $\tilde{z}(d) := 1+ 28/(11d+5)$ and $\varphi(b) := b-\frac12 \log_2{\tilde{z}(C/b)}$.
In our experience, this produces indistinguishable quantization results compared to the more accurate spline construction below. 
Despite the above, we choose to approximate $\varphi$ by a quadratic spline.
With knots $0 = b_0 < b_1 < \dots < b_K = C$
\[
\varphi(b) \;=\; \alpha_k + \beta_k(b - b_k) + \frac{\gamma_k}{2}(b - b_k)^2 \qquad \text{for } b \in [b_k, b_{k+1}] \ .
\]
A spline is more accurate than the above and not hard to obtain. And, it makes it trivial to apply $\varphi$, $\varphi'$ and $\varphi^{-1}$. 
Moreover, conditions on $\varphi$, $\varphi'$, and $\varphi''$ translate to linear constraints on $\alpha$, $\beta$, and $\gamma$. 
This makes the task of finding $\varphi$ a small one-time optimization problem over a convex set of spline parameters.
Note that $\varphi$ depends on the budget $C$.  Below we use $C = 8$ ($k = 256$); splines for other $C$ can be built the same way.

Table~\ref{tab:phi-constraints} lists the constraints that $\varphi$ (and $\alpha$, $\beta$, $\gamma$) should satisfy to realistically model $k$-means. 
For $C = 8$ we use $K = 33$. The knots are $0$, the $17$ measured rates $8/d$ (column $d$ of Table~\ref{tab:quad-spline-knots}), and the geometric midpoints between consecutive rates. Each rate is measured by $k$-means on $\mathcal{N}(0, I_d)$ ($k = 256$; mean of $5$ runs, $1\mathrm{M}$ training samples, evaluated on $200{,}000$ held-out samples).
We impose the coefficient conditions of Table~\ref{tab:phi-constraints}.
The few remaining degrees of freedom are fixed by minimizing least-squares fit over the anchors. 
Table~\ref{tab:quad-spline-knots} lists the resulting coefficients. They satisfy every condition of Table~\ref{tab:phi-constraints}, with $\beta_k \in [0.725, 0.960]$ and $\gamma_k \le 0.42 \cdot 2\ln 2\, \beta_k^2$, and the spline reproduces the measured $z(d)$ to within $0.09\%$, except at $d = 1$, where it uses the exact Lloyd-Max value.

\begin{table}[!htbp]
\caption{Constraints on the effective bit rate $\varphi$ under the separable model $D_j(b_j) = \sigma_j^2\, 2^{-2\varphi(b_j)}$, for any bit budget $C$, and the conditions they imply on the coefficients of the quadratic spline (for all knots $k$). The first row makes $\varphi$ continuously differentiable. Since $\varphi'$ is piecewise linear and $\gamma_k \ge 0$, the conditions at the knots imply the next three constraints on all of $[0, C]$.}
\label{tab:phi-constraints}
\centering
\footnotesize
\renewcommand{\arraystretch}{1.25}
\setlength{\tabcolsep}{4pt}
\begin{tabular}{@{}>{\raggedright\arraybackslash}p{0.22\linewidth}>{\raggedright\arraybackslash}p{0.43\linewidth}>{\raggedright\arraybackslash}p{0.30\linewidth}@{}}
\toprule
Condition & Explanation & Spline coefficient constraints\\
\midrule
$\varphi \in  C^1$ & $\varphi$ is continuous and differentiable & $\alpha_{k+1} = \alpha_k + \beta_k (b_{k+1} - b_k) + \frac{\gamma_k}{2}(b_{k+1} - b_k)^2$ and $\beta_{k+1} = \beta_k + \gamma_k (b_{k+1} - b_k)$ \\
$0 \le \varphi(b) \le b$ & Shannon bound: higher bit allocation never increases distortion, and a codebook of $2^C$ cells carries at most $C$ bits, so $d$ isotropic coordinates sharing it cannot gain more than $C/d$ bits each. In particular $\varphi(0) = 0$: a coordinate without bits costs its full variance. & $\alpha_0 = 0$, $\alpha_k \le b_k$ \\
$0 < \varphi'(b) \le 1$  & Monotonicity and bounded slope. More bits always help, but an extra bit cannot cut the distortion by more than the factor $4$ that it achieves in the high-rate limit. The upper bound is an assumption: it holds at $C = 8$ but fails for very small codebooks (e.g., $C = 1$). & $0 < \beta_k \le 1$ \\
$0 \le \varphi''(b) < 2\ln 2\, \varphi'(b)^2$  & Convexity lower bound: bits are used least efficiently at low rate, where a coordinate is barely resolved, and $\varphi'$ rises toward the high-rate value $1$. Upper bound (equivalently, $2^{-2\varphi}$ is strictly convex): each extra bit reduces the distortion less than the previous one. This makes~\eqref{eq:seg-kkt-def} a convex program whose optimum on an isotropic segment is the uniform allocation. & $0 \le \gamma_k < 2\ln 2\, \beta_k^2$ \\
$\varphi(C/d) \approx C/d - \frac{1}{2}\log_2 z(d)$ & Isotropic agreement: on an isotropic segment, \ez allocates $b_j = C/d$ and reduces to \zg, so it should reproduce the measured $k$-means cost on $\mathcal{N}(0, I_d)$ (Figure~\ref{fig:phi-b}). At $d = 1$ this is exact: the segment is the optimal scalar (Lloyd-Max) quantizer with $2^C$ levels. & $\alpha_k \approx b_k - \frac{1}{2}\log_2 z(d)$ at $b_k = C/d$ \\
\bottomrule
\end{tabular}
\end{table}

\begin{table}[!htbp]
\caption{Coefficients of the quadratic spline $\varphi$ of Appendix~\ref{app:zador-approximations} for $C = 8$. Column $d$ marks the knots $b_k = 8/d$; the remaining knots are geometric midpoints. Six decimals are needed to preserve $C^1$ continuity at the knots.}
\label{tab:quad-spline-knots}
\centering
\footnotesize
\setlength{\tabcolsep}{3.2pt}
\begin{tabular}{@{}rrrrrr@{\hspace{14pt}}rrrrrr@{}}
\toprule
$k$ & $b_k$ & $d$ & $\alpha_k$ & $\beta_k$ & $\gamma_k$ & $k$ & $b_k$ & $d$ & $\alpha_k$ & $\beta_k$ & $\gamma_k$ \\
\midrule
0 & 0.000000 &  & 0.000000 & 0.724877 & 0.302039 & 17 & 0.500000 & 16 & 0.394766 & 0.837669 & 0.094908 \\
1 & 0.031250 & 256 & 0.022800 & 0.734315 & 0.301187 & 18 & 0.577350 &  & 0.459844 & 0.845010 & 0.046303 \\
2 & 0.044194 &  & 0.032330 & 0.738214 & 0.299635 & 19 & 0.666667 & 12 & 0.535501 & 0.849146 & 0.013762 \\
3 & 0.062500 & 128 & 0.045894 & 0.743699 & 0.297110 & 20 & 0.730297 &  & 0.589561 & 0.850022 & 0.001488 \\
4 & 0.072169 &  & 0.053099 & 0.746572 & 0.294042 & 21 & 0.800000 & 10 & 0.648813 & 0.850125 & 0.003950 \\
5 & 0.083333 & 96 & 0.061452 & 0.749855 & 0.290261 & 22 & 0.894427 &  & 0.729106 & 0.850498 & 0.016052 \\
6 & 0.102062 &  & 0.075547 & 0.755291 & 0.285035 & 23 & 1.000000 & 8 & 0.818985 & 0.852193 & 0.027016 \\
7 & 0.125000 & 64 & 0.092947 & 0.761829 & 0.277574 & 24 & 1.154701 &  & 0.951143 & 0.856372 & 0.029834 \\
8 & 0.144338 &  & 0.107730 & 0.767197 & 0.268076 & 25 & 1.333333 & 6 & 1.104595 & 0.861701 & 0.030062 \\
9 & 0.166667 & 48 & 0.124928 & 0.773182 & 0.256378 & 26 & 1.632993 &  & 1.364162 & 0.870710 & 0.031190 \\
10 & 0.204124 &  & 0.154069 & 0.782786 & 0.241595 & 27 & 2.000000 & 4 & 1.685819 & 0.882157 & 0.031004 \\
11 & 0.250000 & 32 & 0.190235 & 0.793869 & 0.224299 & 28 & 2.309401 &  & 1.960243 & 0.891749 & 0.028625 \\
12 & 0.288675 &  & 0.221105 & 0.802544 & 0.205976 & 29 & 2.666667 & 3 & 2.280662 & 0.901976 & 0.023804 \\
13 & 0.333333 & 24 & 0.257151 & 0.811742 & 0.187718 & 30 & 3.265986 &  & 2.825509 & 0.916242 & 0.016337 \\
14 & 0.365148 &  & 0.283071 & 0.817715 & 0.169814 & 31 & 4.000000 & 2 & 3.502444 & 0.928234 & 0.009368 \\
15 & 0.400000 & 20 & 0.311673 & 0.823633 & 0.151912 & 32 & 5.656854 &  & 5.053250 & 0.943755 & 0.006977 \\
16 & 0.447214 &  & 0.350729 & 0.830805 & 0.130030 & 33 & 8.000000 & 1 & 7.283759 & 0.960103 & 0.000000 \\
\bottomrule
\end{tabular}
\end{table}

\FloatBarrier
\section{Properties of \ez}
\label{app:ez-limits}
\begin{theorem}
\label{thm:ez-limits}
The following are properties of \ez:
\begin{enumerate}[label=(\roman*)]
\item Sandwich: $e_{\rwf}^2(g) \le e_{\ez}^2(g)$. If moreover $\varphi(\bar{b}_g) \ge \bar{b}_g - \tfrac12\log_2 z(d_g)$ and $\sigma_j^2 \ge e_{\zg}^2(g)$ for all $j \in g$, then $e_{\ez}^2(g) \le e_{\zg}^2(g)$.
\item High resolution Zador-Gersho limit: for every fixed segment $g$ and every family $\{\varphi_C\}_{C > 0}$ that satisfies (H) below, as $C \to \infty$, $e_{\ez}^2(g)/e_{\zg}^2(g) \to 1$ and all coordinate distortions become equal.
\item Matching \zg on isotropic segments: if $\sigma_j^2 = \sigma^2$ for all $j \in g$, then $b_j^{\ez} = C/d_g$ and $e_{\ez}^2(g) = |\tilde\Sigma_g|^{1/d_g}\, 2^{-2\varphi(\bar{b}_g)}$. In particular, $e_{\ez}^2(g) = e_{\zg}^2(g)$ when $\varphi$ is given by~\eqref{eq:phi-def}.
\item High dimension Shannon limit: let $g^{(m)}$ concatenate $m$ independent copies of $g$ with the corresponding increased budget $mC$. Then, for every family $\{\varphi_C\}_{C > 0}$ that satisfies (H) below, as $m \to \infty$, $e_{\ez}^2(g^{(m)}) / e_{\rwf}^2(g)\rightarrow 1$. Per-coordinate costs converge to the distortion-rate function, including the omission of under-water coordinates.
\item Among active coordinates, larger variance coordinates receive more bits and have \emph{no larger} distortion (strictly smaller when $\varphi$ is strictly convex). Coordinates below the water line are not covered: their distortion $\sigma_j^2$ may be smaller than that of a weakly active coordinate, as in \rwf.
\item Adding coordinates with variance at or below the \ez water line $\theta_{\ez} := \lambda_g/\varphi'(0)$, at a fixed budget $C$, does not move the water line, and leaves the existing allocations and distortions unchanged. 
\end{enumerate}
\end{theorem}

Throughout, let $d = d_g$, $\bar{b}_g = C/d$, $\mathrm{GM} = |\tilde\Sigma_g|^{1/d}$, and $\Delta_j = \tfrac12\log_2(\sigma_j^2/\mathrm{GM})$, so that $\sum_{j\in g} \Delta_j = 0$.
Recall from Section~\ref{sec:ezg} that $\varphi$ satisfies (a) $\varphi(0) = 0$, (b) $0 < \varphi'(b) \le 1$, and (c) $0 \le \varphi''(b) < (2\ln 2)\varphi'(b)^2$.
In particular, $D_j(b_j) = \sigma_j^2 2^{-2\varphi(b_j)}$ is strictly decreasing and strictly convex, so the \ez minimizer $b^{\ez}$ in~\eqref{eq:seg-kkt-def} is unique, satisfies $\sum_{j\in g} b_j^{\ez} = C$, and is characterized by the KKT conditions: there exists $\lambda_g > 0$ such that
\begin{equation}
\label{eq:app-ez-kkt}
D_j(b_j^{\ez})\,\varphi'(b_j^{\ez}) = \lambda_g \quad \text{if } b_j^{\ez} > 0,
\qquad
\sigma_j^2\,\varphi'(0) \le \lambda_g \quad \text{if } b_j^{\ez} = 0.
\end{equation}
Parts (ii) and (iv) vary the budget, so there $\varphi = \varphi_C$ is a family indexed by $C$, each member satisfying (a)--(c) on $[0, C]$. Every such $\varphi_C$ can be written in the form~\eqref{eq:phi-def} with its own finite-rate factor $z_C(t) := 2^{2(C/t - \varphi_C(C/t))}$, $t \ge 1$, in place of $z$. When $\varphi = \varphi_C$ is given by~\eqref{eq:phi-def} in this sense, i.e.\ $\varphi_C(b) = b - \tfrac12\log_2 z_C(C/b)$, we have $D_j(b_j) = \sigma_j^2 2^{-2b_j} z_C(C/b_j)$ for $b_j > 0$ and $D_j(0) = \sigma_j^2$.
By (a) and (b), $b - \varphi_C(b) = \tfrac12\log_2 z_C(C/b)$ vanishes at $b=0$ and has derivative $1 - \varphi_C'(b) \ge 0$, so $z_C$ is non-increasing with $1 \le z_C(t) \le z_C(1)$, continuous on $[1,\infty)$, and satisfies $z_C(t) \to 1$ as $t \to \infty$. The factor $z_C$ is a finite-rate quantity and in general differs from the classical constant $z$ of Section~\ref{sec:zg}. In fact $z$ itself cannot be used in~\eqref{eq:phi-def} at a finite budget: by the sphere lower bound on $c_t$, $z(t) \ge 1 + (\ln(\pi t) - 1)/t + O(\ln^2 t / t^2)$, so $C/t - \tfrac12\log_2 z(t)$ is negative for all sufficiently large integers $t = C/b$, which contradicts (a)--(b). Letting $b \to 0^+$ at a fixed $C$ probes the low-rate regime, not the high-rate one. The two limits therefore need a condition on the family. Parts (ii) and (iv) assume that the family approaches the Zador-Gersho rate penalty uniformly as the budget grows:
(H) there is a continuous function $\bar z$ on $[1, \infty)$ with $\bar z(d) = z(d)$ at the integers $d$ and $\bar z(t) \to 1$ as $t \to \infty$, such that $\sup_{t \ge 1} |z_C(t) - \bar z(t)| \to 0$ as $C \to \infty$.
Equivalently, $\varphi_C(b) = b - \tfrac12\log_2 \bar z(C/b) + o(1)$ uniformly in $b \in (0, C]$: at large budgets, the effective rate is the Zador-Gersho rate at every segment width. Under (H), $z_C(t_C) \to z(d_g)$ whenever $t_C \to d_g$ as $C \to \infty$, by continuity of $\bar z$, and $z_{mC}(m) \to 1$ as $m \to \infty$, since $|z_{mC}(m) - \bar z(m)| \to 0$ and $\bar z(m) \to 1$. Moreover, $z_C \le z_C(1) \to z(1)$, so $Z := \sup_{C \ge C_0} z_C(1) < \infty$ for some $C_0$; since parts (ii) and (iv) concern $C \to \infty$, we take $C \ge C_0$ there.

\paragraph{Proof of (i): sandwiching between \rwf and \zg.}
By (a) and (b), $\varphi(b_j) = \int_0^{b_j} \varphi'(t)\,dt \le b_j$ for all $b_j \ge 0$.
Since $b^{\ez} \ge 0$ and $\sum_{j\in g} b_j^{\ez} = C$, $b^{\ez}$ is feasible for the \rwf program~\eqref{eq:rwf-program}, giving
\[
d\, e_{\ez}^2(g)
= \sum_{j\in g} \sigma_j^2\, 2^{-2\varphi(b_j^{\ez})}
\ge \sum_{j\in g} \sigma_j^2\, 2^{-2b_j^{\ez}}
\ge d\, e_{\rwf}^2(g).
\]
For the upper bound, assume $\sigma_j^2 \ge e_{\zg}^2(g) \ge \mathrm{GM}\, 2^{-2\varphi(\bar{b}_g)}$ for all $j\in g$, and define
\[
\tilde{b}_j := \varphi^{-1}\!\Big(\varphi(\bar{b}_g) + \tfrac12\log_2\frac{\sigma_j^2}{\mathrm{GM}}\Big).
\]
Each $\tilde{b}_j \ge 0$ because $\varphi(\bar{b}_g) + \tfrac12\log_2(\sigma_j^2/\mathrm{GM}) \ge \tfrac12\log_2(\sigma_j^2 / e_{\zg}^2(g)) \ge 0$.
Moreover, writing $y_j := \varphi(\bar{b}_g) + \tfrac12\log_2(\sigma_j^2/\mathrm{GM}) \ge 0$, since $\sum_{k \in g} y_k = d\,\varphi(C/d)$ and $\varphi$ is convex on $[0, C]$ with $\varphi(0) = 0$, each $y_j = d\,\varphi(C/d) - \sum_{k \ne j} y_k \le d\,\varphi(C/d) \le \varphi(C)$, so $y_j \in [0, \varphi(C)]$ and $\tilde{b}_j = \varphi^{-1}(y_j)$ is well-defined.
Moreover, since $\varphi$ is strictly increasing and convex by (b)-(c), its inverse $\varphi^{-1}$ is concave. By Jensen's inequality and $\sum_{j\in g} \log_2(\sigma_j^2/\mathrm{GM}) = 0$,
\[
\frac{1}{d}\sum_{j\in g} \tilde{b}_j
\le \varphi^{-1}\!\Big(\varphi(\bar{b}_g) + \frac{1}{2d}\sum_{j\in g} \log_2\frac{\sigma_j^2}{\mathrm{GM}}\Big)
= \bar{b}_g,
\]
so $\sum_{j\in g} \tilde{b}_j \le C$. Hence $\tilde{b}$ is feasible for~\eqref{eq:seg-kkt-def}, and by definition $D_j(\tilde{b}_j) = \sigma_j^2 2^{-2\varphi(\tilde{b}_j)} = \mathrm{GM}\, 2^{-2\varphi(\bar{b}_g)} \le e_{\zg}^2(g)$ for every $j \in g$. Thus $e_{\ez}^2(g) \le \frac{1}{d}\sum_{j\in g} D_j(\tilde{b}_j) \le e_{\zg}^2(g)$. \qed

The condition $\varphi(\bar{b}_g) \ge \bar{b}_g - \tfrac12\log_2 z(d_g)$ says that \ez's cost on an isotropic segment is at most the high-rate \zg cost. It cannot be dropped: on an isotropic segment, part (iii) gives $e_{\ez}^2(g) = \mathrm{GM}\,2^{-2\varphi(\bar{b}_g)}$, so $e_{\ez}^2(g) \le e_{\zg}^2(g)$ holds if and only if the condition holds. It is automatic when $\varphi$ is given by~\eqref{eq:phi-def}, and it holds for the spline of Appendix~\ref{app:zador-approximations} ($C = 8$) at every width $d_g \le 2048$, by a margin of at least $0.001$ bits. For this check we used the exact $c_1$ and $c_2$, and for $d \ge 3$ the sphere lower bound on $c_d$, which can only make $z(d)$ smaller.

\paragraph{Proof of (ii): high-resolution \zg limit.}
Let $V(C) = d\, e_{\ez}^2(g)$ and $Z_g(C) = d\, e_{\zg}^2(g) = d\, z(d)\, \mathrm{GM}\, 2^{-2C/d}$.
Since $1 \le z_C \le Z$, every $b_j \ge 0$ satisfies
\begin{equation}
\label{eq:app-lim-sandwich}
\sigma_j^2\, 2^{-2b_j} \;\le\; D_j(b_j) \;\le\; Z\, \sigma_j^2\, 2^{-2b_j}.
\end{equation}
\emph{Upper bound.} For $C/d > \max_j |\Delta_j|$, the Shannon allocation $\hat{b}_j = C/d + \Delta_j$ is positive and satisfies $\sigma_j^2 2^{-2\hat{b}_j} = \mathrm{GM}\, 2^{-2C/d}$ for every $j$. Hence
\[
V(C) \;\le\; \mathrm{GM}\, 2^{-2C/d} \sum_{j\in g} z_C(C/\hat{b}_j).
\]
Since $C/\hat{b}_j = d/(1 + d\Delta_j/C) \to d$, (H) gives $z_C(C/\hat{b}_j) \to z(d)$, and the right-hand side is $Z_g(C)(1 + o(1))$. Bounding $z_C \le Z$ also gives $V(C) \le dZ\,\mathrm{GM}\,2^{-2C/d}$.

\emph{Localization.} For the minimizer $b = b^{\ez}$,~\eqref{eq:app-lim-sandwich} and the crude upper bound give $\sigma_j^2 2^{-2b_j} \le D_j(b_j) \le V(C) \le dZ\,\mathrm{GM}\,2^{-2C/d}$, so $b_j \ge C/d + \Delta_j - \mu$ with $\mu = \tfrac12\log_2(dZ)$. Since $\sum_i b_i = C$, $b_j = C - \sum_{i\ne j} b_i \le C/d + \Delta_j + (d-1)\mu$. Thus
\begin{equation}
\label{eq:app-lim-band}
\Delta_j - \mu \;\le\; b_j - C/d \;\le\; \Delta_j + (d-1)\mu,
\end{equation}
so every coordinate is active for large $C$ and $C/b_j \to d$ for all $j$.

\emph{Lower bound and equal distortions.} By~\eqref{eq:app-lim-band} and (H), $z_C(C/b_j) \ge z(d)(1 - \varepsilon_C)$ with $\varepsilon_C \to 0$. Setting $x_j = \sigma_j^2 2^{-2b_j}$ (so $\prod_j x_j = \mathrm{GM}^d 2^{-2C}$), AM-GM gives
\[
V(C) = \sum_{j\in g} x_j\, z_C(C/b_j) \ge z(d)(1-\varepsilon_C) \sum_{j\in g} x_j \ge (1-\varepsilon_C) Z_g(C).
\]
Combined with the upper bound, $V(C)/Z_g(C) \to 1$, so $e_{\ez}^2(g)/e_{\zg}^2(g) \to 1$.
Finally, let $y_j = x_j/(\mathrm{GM}\,2^{-2C/d})$, so $\prod_j y_j = 1$ and $\sum_j y_j \le d(1 + \eta_C)$ with $\eta_C \to 0$. Writing $t_j = \ln y_j$ gives $\sum_j t_j = 0$ and $\sum_j (e^{t_j} - 1 - t_j) = \sum_j y_j - d \le d\eta_C$. Since $e^t - 1 - t \ge 0$ vanishes only at $t = 0$, $t_j \to 0$ and $y_j \to 1$, whence $D_j(b_j) = \mathrm{GM}\,2^{-2C/d}\, y_j\, z_C(C/b_j) = e_{\zg}^2(g)(1 + o(1))$ for every $j$. \qed

\paragraph{Proof of (iii): matching \zg on isotropic segments.}
When $\sigma_j^2 = \sigma^2$ for all $j \in g$, all $D_j(b_j) = \sigma^2 2^{-2\varphi(b_j)}$ are identical strictly convex functions. By Jensen's inequality (or symmetry), the unique minimizer of $\sum_{j\in g} D_j(b_j)$ subject to $\sum_{j\in g} b_j \le C$ is $b_j^{\ez} = C/d = \bar{b}_g$ for all $j\in g$. Thus $e_{\ez}^2(g) = \sigma^2 2^{-2\varphi(\bar{b}_g)} = |\tilde\Sigma_g|^{1/d} 2^{-2\varphi(\bar{b}_g)}$, which equals $e_{\zg}^2(g)$ when $\varphi$ is given by~\eqref{eq:phi-def}. \qed

\paragraph{Proof of (iv): high-dimension Shannon limit.}
Let $g^{(m)}$ concatenate $m$ copies of $g$ with budget $mC$ (so $\varphi = \varphi_{mC}$ and the factor is $z_{mC}$). By scaling~\eqref{eq:rwf-program}, $e_{\rwf}^2(g^{(m)}) = e_{\rwf}^2(g)$, and part (i) gives $e_{\ez}^2(g^{(m)}) \ge e_{\rwf}^2(g)$.
For the upper bound, replicating the \rwf allocation $b^{\rwf}$ across all $m$ copies is feasible for $g^{(m)}$. Each active coordinate has $b_j^{\rwf} \le C$, so $mC/b_j^{\rwf} \ge m$ and $z_{mC}(mC/b_j^{\rwf}) \le z_{mC}(m)$ because $z_{mC}$ is non-increasing (while $D_j(0) = \sigma_j^2 \le z_{mC}(m)\sigma_j^2$). Summing over all $md_g$ coordinates yields $e_{\ez}^2(g^{(m)}) \le z_{mC}(m)\, e_{\rwf}^2(g)$. Since $z_{mC}(m) \to 1$ as $m \to \infty$ by (H), $e_{\ez}^2(g^{(m)})/e_{\rwf}^2(g) \to 1$.

For per-coordinate convergence, symmetry and strict convexity imply that all $m$ copies of coordinate $j \in g$ receive the same \ez allocation $b_j^{(m)}$ with $\sum_{j\in g} b_j^{(m)} = C$. Since $\varphi_{mC}(b) \le b$,
\[
e_{\rwf}^2(g) \le \frac{1}{d_g} \sum_{j\in g} \sigma_j^2\, 2^{-2b_j^{(m)}} \le e_{\ez}^2(g^{(m)}) \le z_{mC}(m)\, e_{\rwf}^2(g) \to e_{\rwf}^2(g).
\]
By compactness of $\{b \ge 0 : \sum_{j\in g} b_j = C\}$ and uniqueness of the strictly convex \rwf minimizer $b^{\rwf}$, $b_j^{(m)} \to b_j^{\rwf}$ for every $j \in g$.
If $b_j^{\rwf} = 0$ (under-water), then $\sigma_j^2 2^{-2b_j^{(m)}} \le D_j^{(m)}(b_j^{(m)}) \le \sigma_j^2$ gives $D_j^{(m)}(b_j^{(m)}) \to \sigma_j^2$.
If $b_j^{\rwf} > 0$, then $mC/b_j^{(m)} \ge m$, so $1 \le z_{mC}(mC/b_j^{(m)}) \le z_{mC}(m) \to 1$ and $D_j^{(m)}(b_j^{(m)}) = \sigma_j^2 2^{-2b_j^{(m)}} z_{mC}(mC/b_j^{(m)}) \to \sigma_j^2 2^{-2b_j^{\rwf}}$. \qed

\paragraph{Proof of (v): among active coordinates, larger variance gives no larger distortion.}
Let $i, j \in g$ have $\sigma_i^2 > \sigma_j^2$, with both coordinates active ($b_i^{\ez}, b_j^{\ez} > 0$).
Note that $D_k(b)\varphi'(b) = \sigma_k^2 \cdot \big(-\tfrac{1}{2\ln 2}\tfrac{d}{db} 2^{-2\varphi(b)}\big)$, and strict convexity of $2^{-2\varphi(b)}$ implies $b \mapsto 2^{-2\varphi(b)}\varphi'(b)$ is strictly decreasing.
By~\eqref{eq:app-ez-kkt}, $\sigma_i^2 2^{-2\varphi(b_i^{\ez})}\varphi'(b_i^{\ez}) = \lambda_g = \sigma_j^2 2^{-2\varphi(b_j^{\ez})}\varphi'(b_j^{\ez})$; since $\sigma_i^2 > \sigma_j^2$, this forces $b_i^{\ez} > b_j^{\ez}$.
Since $\varphi$ is convex by (c), $\varphi'$ is non-decreasing, so $\varphi'(b_i^{\ez}) \ge \varphi'(b_j^{\ez}) > 0$. Using~\eqref{eq:app-ez-kkt} again,
\[
D_i(b_i^{\ez})
= \frac{\lambda_g}{\varphi'(b_i^{\ez})}
\le \frac{\lambda_g}{\varphi'(b_j^{\ez})}
= D_j(b_j^{\ez}),
\]
with strict inequality whenever $\varphi'$ is strictly increasing on $[b_j^{\ez}, b_i^{\ez}]$. \qed

The restriction to active coordinates is necessary. If $j$ is inactive with $\sigma_j^2$ just below $\theta_{\ez} = \lambda_g/\varphi'(0)$ and $i$ is active with $b_i^{\ez}$ close to $0$, then $D_j = \sigma_j^2 < \sigma_i^2 \approx D_i(b_i^{\ez})$, as in \rwf, where active coordinates sit at the water level and inactive ones below it.

\paragraph{Proof of (vi): adding coordinates below the water line.}
By~\eqref{eq:app-ez-kkt}, the \ez water line is $\theta_{\ez} := \lambda_g / \varphi'(0)$: a coordinate $j$ is active ($b_j^{\ez} > 0$) if and only if $\sigma_j^2 > \theta_{\ez}$.
Suppose we append coordinates $k$ with $\sigma_k^2 \le \theta_{\ez}$ to $g$ while keeping the budget $C$ fixed, and extend $b^{\ez}$ by setting $\tilde{b}_k = 0$ for each new coordinate.
Then $\tilde{b}$ remains feasible ($\sum \tilde{b}_j = C$, $\tilde{b} \ge 0$), all existing coordinates still satisfy~\eqref{eq:app-ez-kkt} with the same $\lambda_g$, and each new coordinate satisfies $\sigma_k^2 \varphi'(0) \le \theta_{\ez} \varphi'(0) = \lambda_g$.
Since~\eqref{eq:seg-kkt-def} is convex, the KKT conditions are sufficient for optimality; thus $\tilde{b}$ is optimal, the water line $\theta_{\ez}$ does not move, and existing allocations and distortions are unchanged. \qed


\FloatBarrier
\section{Extended Zador Estimator (\ez) Empirical Evaluation}
\label{app:coupled-zador-validation}
We evaluate \rwf~\eqref{eq:rwf}, \zg~\eqref{eqn:zador}, and \ez~\eqref{eq:seg-kkt-def} with the spline $\varphi$ of Table~\ref{tab:quad-spline-knots} compared to actual $k$-means experimental runs. Here \zg is evaluated at finite rate, as $|\tilde{\Sigma}_g|^{1/d}\, 2^{-2\varphi(C/d)}$ with the spline $\varphi$ (the form~\eqref{eq:phi-def}), not with the $C \to \infty$ constant $z(d)$ of~\eqref{eqn:zador}. The comparison therefore isolates the error caused by the skew of the spectrum, after isotropic calibration. 
We test $8$ diverse spectral shapes: log-linear, Zipf power-law, convex quadratic scree, concave tail cliff, bimodal $50/50$ step, dominant top-$1/4$ block, steep cubic scree, and cosine S-curve.
Each is configured with a skewness parameter $\alpha \in \{0, 0.25, 0.5, 0.75, 1.0\}$, where $\alpha=0$ means isotropy and $\alpha=1.0$ is the most skewed configuration. 
We repeat this four times for $d \in \{4, 8, 16, 32\}$. 
This gives $160$ configurations ($128$ non-isotropic with $\alpha > 0$).

These are compared to real $k$-means runs.
The clusters were computed using $1{,}000{,}000$ randomly generated Gaussians with variances specified by the relevant spectra. 
The clustering cost was measured on a held out set of $200{,}000$ fresh test samples. 
The experiment was run $5$ times and the median is reported. 

Figure~\ref{fig:crwf-validation} plots the ratio $e_{\mathrm{est}}^2 / e_{\mathrm{actual}}^2$ across all $32$ dimension-spectrum curves, and Table~\ref{tab:crwf-validation} summarizes the error statistics.
Since \rwf is a lower bound it unsurprisingly underestimates everywhere. 
As predicted by theory, the higher the dimension, the less significant the underestimation error is. 
Again, in accordance with theory, \zg is exact at $\alpha = 0$ (by construction: the isotropic configurations are among the measurements that $\varphi$ is fit to) but overestimates quite significantly as the skew grows. 
Finally, \ez almost exactly removes the fan-out. Evaluating $\varphi(b_j)$ at each coordinate's own bit allocation reduces the non-isotropic error to $0.51\%$ MAPE ($0.80\%$ RMS, $+0.48\%$ bias, and $2.30\%$ max error; $0.41\%$ MAPE over all $160$ configurations).

Note that at $d = 4$ and $d = 8$, the \ez ratio rises slightly with skewness (at most $+2.30\%$ and $+1.50\%$, respectively), whereas at $d = 32$ it remains flat within $[-0.12\%, +0.19\%]$. A finite codebook of $k = 256$ centroids cannot realize a fractional number of quantization levels along an axis. We conjecture that $k$-means exhibits discrete level effects at small $d$ that no separable continuous $\varphi$ can capture.

\begin{table}[!htbp]
\caption{Detailed error breakdown (\%) of \rwf~\eqref{eq:rwf}, \zg~(\ref{eq:phi-def}), and \ez~\eqref{eq:seg-kkt-def} against $1\mathrm{M}$-sample $k$-means ($k = 256, C = 8$) using the spline $\varphi$ of Table~\ref{tab:quad-spline-knots}: (a) aggregate metrics over the $128$ non-isotropic configurations ($\alpha > 0$) and MAPE at each skewness $\alpha \in \{0, 0.25, 0.5, 0.75, 1.0\}$; (b) non-isotropic MAPE, $\alpha = 1$ MAPE, and maximum absolute error by dimension $d \in \{4, 8, 16, 32\}$; and (c) non-isotropic MAPE by spectral shape.}
\label{tab:crwf-validation}
\centering
\footnotesize
\setlength{\tabcolsep}{4.2pt}

\medskip
\textbf{(a) Aggregate Non-Isotropic Error ($\alpha > 0$) and MAPE by Skewness $\alpha$} \\[3pt]
\begin{tabular}{@{}lccccccccc@{}}
\toprule
Estimator & MAPE & RMS & Bias & Max $|\text{Err}|$ & $\alpha{=}0$ & $\alpha{=}0.25$ & $\alpha{=}0.50$ & $\alpha{=}0.75$ & $\alpha{=}1.00$ \\
\midrule
\rwf~\eqref{eq:rwf}               & $18.41\%$ & $20.72\%$ & $-18.41\%$ & $35.17\%$ & $19.76\%$ & $19.56\%$ & $18.95\%$ & $18.01\%$ & $17.13\%$ \\
\zg~(\ref{eq:phi-def})             & $1.87\%$  & $3.13\%$  & $+1.87\%$  & $14.22\%$ & $\mathbf{0.00\%}$ & $0.28\%$  & $1.12\%$  & $2.43\%$  & $3.63\%$  \\
\textbf{\ez}~\eqref{eq:seg-kkt-def} & $\mathbf{0.51\%}$ & $\mathbf{0.80\%}$ & $\mathbf{+0.48\%}$ & $\mathbf{2.30\%}$ & $\mathbf{0.00\%}$ & $\mathbf{0.11\%}$ & $\mathbf{0.43\%}$ & $\mathbf{0.85\%}$ & $\mathbf{0.66\%}$ \\
\bottomrule
\end{tabular}

\vspace{8pt}
\textbf{(b) Non-Isotropic MAPE, $\alpha = 1$ MAPE, and Max $|\text{Err}|$ by Dimension $d$} \\[3pt]
\setlength{\tabcolsep}{3.2pt}
\begin{tabular}{@{}lcccccccccccc@{}}
\toprule
 & \multicolumn{3}{c}{$d = 4$} & \multicolumn{3}{c}{$d = 8$} & \multicolumn{3}{c}{$d = 16$} & \multicolumn{3}{c}{$d = 32$} \\
\cmidrule(lr){2-4} \cmidrule(lr){5-7} \cmidrule(lr){8-10} \cmidrule(l){11-13}
Estimator & MAPE & $\alpha{=}1$ & Max & MAPE & $\alpha{=}1$ & Max & MAPE & $\alpha{=}1$ & Max & MAPE & $\alpha{=}1$ & Max \\
\midrule
\rwf~\eqref{eq:rwf}               & $32.47\%$ & $29.94\%$ & $35.17\%$ & $20.83\%$ & $19.45\%$ & $22.12\%$ & $12.79\%$ & $11.97\%$ & $13.55\%$ & $7.55\%$ & $7.17\%$ & $7.93\%$ \\
\zg~(\ref{eq:phi-def})             & $4.38\%$  & $8.30\%$  & $14.22\%$ & $1.75\%$  & $3.53\%$  & $7.35\%$  & $0.91\%$  & $1.86\%$  & $3.79\%$  & $0.43\%$ & $0.84\%$ & $2.06\%$ \\
\textbf{\ez}~\eqref{eq:seg-kkt-def} & $\mathbf{1.15\%}$ & $\mathbf{1.46\%}$ & $\mathbf{2.30\%}$ & $\mathbf{0.69\%}$ & $\mathbf{0.89\%}$ & $\mathbf{1.50\%}$ & $\mathbf{0.16\%}$ & $\mathbf{0.24\%}$ & $\mathbf{0.83\%}$ & $\mathbf{0.05\%}$ & $\mathbf{0.07\%}$ & $\mathbf{0.19\%}$ \\
\bottomrule
\end{tabular}

\vspace{8pt}
\textbf{(c) Non-Isotropic MAPE ($\alpha > 0$) by Spectral Shape} \\[3pt]
\setlength{\tabcolsep}{4.0pt}
\begin{tabular}{@{}lcccccccc@{}}
\toprule
Estimator & Log-Linear & Zipf & Quad Scree & Tail Cliff & 50/50 Step & Top-$1/4$ & Cubic Scree & Cosine \\
\midrule
\rwf~\eqref{eq:rwf}               & $18.96\%$ & $18.64\%$ & $18.42\%$ & $19.18\%$ & $18.27\%$ & $17.02\%$ & $18.02\%$ & $18.79\%$ \\
\zg~(\ref{eq:phi-def})             & $1.12\%$  & $1.55\%$  & $1.87\%$  & $0.83\%$  & $2.06\%$  & $3.75\%$  & $2.40\%$  & $1.36\%$  \\
\textbf{\ez}~\eqref{eq:seg-kkt-def} & $\mathbf{0.40\%}$ & $\mathbf{0.52\%}$ & $\mathbf{0.62\%}$ & $\mathbf{0.28\%}$ & $\mathbf{0.55\%}$ & $\mathbf{0.60\%}$ & $\mathbf{0.67\%}$ & $\mathbf{0.48\%}$ \\
\bottomrule
\end{tabular}
\end{table}

\FloatBarrier
\section{Pessimality of Random Rotation}
\label{app:random-rotation}
\label{app:proofs}

Theorem~\ref{thm:pca} proves that the SVD rotation is optimal for exact Gaussian $k$-means PQ (and under \ez). Here, we show that random rotations approximate the \emph{worst} variance profile for the diagonal cost $f_G$ introduced in the proof.

\begin{theorem}[Pessimality of the Uniform Diagonal]
\label{thm:uniform-diagonal}
Let $\Sigma$ with eigenvalues $\sigma_j^2$ be as in Theorem~\ref{thm:pca}, and set
\[
\bar s^2 := (\frac{1}{d}\sum_{j=1}^d \sigma_j^2) \mathbf{1},
\qquad
\text{then}
\qquad
\bar s^2 \in \arg \max_{s^2 \in \mathcal{M}(\sigma^2)} \left[\min_G f_G(s^2)\right] 
\qquad
\text{and}
\qquad
\bar s^2 = \diag(R_\star\Sigma R_\star^\top)
\]
for some rotation $R_\star$ where $f_G$ is the diagonal cost from~\eqref{eq:mse-s2-decomp}.
\end{theorem}

\begin{proof}
By the proof of Theorem~\ref{thm:pca}, $\min_G f_G(s^2)$ is concave and permutation-symmetric on $\mathcal{M}(\sigma^2)$.
Write $h(u) := \min_G f_G(u)$. For every $u \in \mathcal{M}(\sigma^2)$, the average of its permutations is $\frac{1}{d!}\sum_{P\in\Pi_d} Pu = \bar{s}^2$, because all points of $\mathcal{M}(\sigma^2)$ have the same coordinate sum. By concavity (Jensen's inequality) and symmetry, $h(\bar{s}^2) \ge \frac{1}{d!}\sum_{P\in\Pi_d} h(Pu) = h(u)$. Hence $\bar{s}^2$ maximizes $h$ over $\mathcal{M}(\sigma^2)$; it need not be the only maximizer, since $h$ is not strictly concave in general. 
Since $\bar{s}^2 = \frac{1}{d!}\sum_{P\in\Pi_d}P\sigma^2 \in \mathcal{M}(\sigma^2)$, Schur-Horn guarantees a rotation $R_\star \in O(d)$ with $\diag(R_\star\Sigma R_\star^\top) = \bar{s}^2$.
\end{proof}

We note that uniform random rotations yield covariance matrices that tightly concentrate around the flat spectrum.
We state (without providing the formal proof) that a typical random rotation therefore results in near \emph{maximal} distortion of PQ.

\FloatBarrier

\section{Recall on \vqb}
\label{app:vqbench-recall}

This subsection supplements Section~\ref{sec:vq-bench-results} with the recall@$10$ of the quantizers of Table~\ref{tab:vqbench-recon} (Table~\ref{tab:vqbench-recall}). Recall is measured by re-ranking: for each query, the top $10$ of its $100$ candidates under the quantized inner products is compared with their top $10$ under the exact inner products. Vectors outside this pool are never scored, so these numbers are not full-corpus retrieval recalls. \ezpq has a higher recall than every quantizer outside the PQ family on $38$ of the $40$ dataset-rate pairs. The exceptions are Yahoo at $\bar{b} = 0.5$ ($0.454$ versus $0.459$ for QJL) and at $\bar{b} = 4$ ($0.936$ versus $0.943$ for EDEN-prod); on MS MARCO at $\bar{b} = 4$ the two agree to three decimals ($0.965$). These quantizers use up to $0.75$ more code bits per dimension than \ezpq (Section~\ref{sec:vq-bench-results}), and QJL and EDEN-prod are designed to estimate inner products rather than to minimize the reconstruction error. Within the PQ family the differences are small: \ezpq's recall is within $0.02$ of OPQ's and of OPQ-P's on every dataset and rate, and at $\bar{b} = 0.5$ the better of OPQ and OPQ-P exceeds \ezpq on $8$ of the $10$ datasets.

\begin{table}[!htbp]
\caption{Recall@$10$ on the ten \vqb datasets at $\bar{b}$ code bits per dimension (excluding model overhead). \emph{Best other} is the highest recall among the $12$ quantizers outside the PQ family at the same nominal bit width: QJL at $\bar{b} = 0.5$; EDEN-prod, ITQ-asym, RaBitQ, Scalar or TurboQuant-MSE at $\bar{b} = 1$; EDEN-prod, EDEN-MSE or E-RaBitQ at $\bar{b} = 2$; and EDEN-prod, EDEN-MSE or Scalar at $\bar{b} = 4$. Bold: best per dataset and rate.}
\label{tab:vqbench-recall}
\centering
\footnotesize
\setlength{\tabcolsep}{3.5pt}
\resizebox{\linewidth}{!}{%
\begin{tabular}{@{}llrrrrrrrrrr@{}}
\toprule
$\bar{b}$ & \textbf{Method} & \textbf{Llama} & \textbf{Yahoo} & \textbf{ImageNet} & \textbf{COCO} & \textbf{CCNews} & \textbf{ArXiv} & \textbf{HotpotQA} & \textbf{LAION} & \textbf{Landmark} & \textbf{MS MARCO} \\
 & & $d = 128$ & $d = 384$ & $d = 512$ & $d = 768$ & $d = 768$ & $d = 768$ & $d = 640$ & $d = 512$ & $d = 768$ & $d = 1024$ \\
\midrule
$0.5$ & \ezpq & 0.200 & 0.454 & 0.354 & 0.326 & \textbf{0.632} & 0.537 & 0.443 & 0.276 & 0.418 & \textbf{0.681} \\
 & OPQ-P & \textbf{0.204} & 0.448 & 0.346 & \textbf{0.332} & 0.625 & 0.536 & 0.443 & 0.277 & 0.418 & 0.679 \\
 & OPQ & 0.188 & \textbf{0.461} & \textbf{0.360} & 0.325 & 0.625 & \textbf{0.543} & \textbf{0.451} & \textbf{0.289} & \textbf{0.421} & \textbf{0.681} \\
 & PQ & 0.136 & 0.442 & 0.245 & 0.150 & 0.493 & 0.376 & 0.402 & 0.233 & 0.259 & 0.644 \\
 & Best other & 0.128 & 0.459 & 0.322 & 0.162 & 0.508 & 0.410 & 0.364 & 0.230 & 0.303 & 0.670 \\
\midrule
$1$ & \ezpq & 0.247 & \textbf{0.653} & \textbf{0.548} & \textbf{0.488} & \textbf{0.773} & \textbf{0.728} & 0.621 & 0.446 & \textbf{0.616} & \textbf{0.818} \\
 & OPQ-P & 0.262 & 0.648 & 0.544 & 0.485 & 0.766 & 0.724 & 0.619 & \textbf{0.451} & 0.612 & 0.813 \\
 & OPQ & \textbf{0.264} & 0.652 & 0.547 & 0.479 & 0.772 & 0.721 & \textbf{0.626} & 0.446 & 0.615 & 0.813 \\
 & PQ & 0.141 & 0.620 & 0.397 & 0.211 & 0.636 & 0.551 & 0.551 & 0.372 & 0.384 & 0.753 \\
 & Best other & 0.191 & 0.634 & 0.466 & 0.243 & 0.647 & 0.576 & 0.540 & 0.365 & 0.428 & 0.788 \\
\midrule
$2$ & \ezpq & \textbf{0.371} & \textbf{0.826} & \textbf{0.753} & \textbf{0.688} & \textbf{0.888} & \textbf{0.876} & 0.792 & 0.688 & \textbf{0.791} & \textbf{0.908} \\
 & OPQ-P & 0.365 & \textbf{0.826} & 0.752 & 0.682 & 0.887 & 0.868 & \textbf{0.795} & 0.689 & 0.789 & 0.907 \\
 & OPQ & 0.369 & 0.824 & \textbf{0.753} & 0.681 & 0.885 & 0.870 & 0.792 & \textbf{0.696} & 0.783 & 0.907 \\
 & PQ & 0.225 & 0.783 & 0.618 & 0.390 & 0.791 & 0.755 & 0.743 & 0.604 & 0.618 & 0.861 \\
 & Best other & 0.222 & 0.808 & 0.677 & 0.385 & 0.801 & 0.769 & 0.730 & 0.543 & 0.633 & 0.884 \\
\midrule
$4$ & \ezpq & 0.645 & 0.936 & 0.900 & \textbf{0.860} & 0.949 & \textbf{0.946} & \textbf{0.927} & 0.888 & \textbf{0.901} & \textbf{0.965} \\
 & OPQ-P & 0.645 & 0.936 & 0.900 & 0.858 & \textbf{0.951} & \textbf{0.946} & \textbf{0.927} & \textbf{0.890} & 0.900 & 0.964 \\
 & OPQ & \textbf{0.658} & 0.937 & \textbf{0.902} & 0.853 & 0.950 & 0.945 & 0.922 & 0.888 & 0.897 & 0.963 \\
 & PQ & 0.528 & 0.921 & 0.841 & 0.747 & 0.924 & 0.918 & 0.909 & 0.868 & 0.861 & 0.948 \\
 & Best other & 0.549 & \textbf{0.943} & 0.893 & 0.755 & 0.936 & 0.928 & 0.917 & 0.854 & 0.876 & \textbf{0.965} \\
\bottomrule
\end{tabular}}
\end{table}
\FloatBarrier

\section{Timing Results on \vqb}
\label{app:vqbench-timing}

This subsection supplements Section~\ref{sec:vq-bench-results} with the encode time (Table~\ref{tab:vqbench-encode}) and scoring time (Table~\ref{tab:vqbench-score}) of the PQ-family quantizers on each dataset. Encode time is the wall-clock time to read and encode the whole database; scoring time is the median time to score one query against its $100$ candidates. All timings were measured on a $64$-core AMD EPYC 7B13 machine ($120$\,GB RAM), and every quantizer encoded with all $64$ threads. With only $100$ candidates, fewer than the $k = 256$ centroids of a segment, the scoring time is dominated by the per-query setup, i.e., the $O(dd')$ projection and the $O(d'k)$ construction of the lookup tables, and not by the $O(m)$ table lookups per candidate. In a full scan or an IVF probe over $10^4$ to $10^6$ candidates the setup is amortized, and the per-candidate cost, which is the same for all PQ-family methods at equal $m$, dominates. Fit times (on the $25{,}000$ training vectors) are not tabulated. Over all datasets and rates, \ezpq's fit takes under $43$ seconds and between $0.6$ and $1.2$ times OPQ-P's, the alternating optimization of OPQ takes $7$ to $16$ times longer than \ezpq, and the random-rotation quantizers (EDEN, RaBitQ, E-RaBitQ, TurboQuant) fit in at most $7$ seconds.

\begin{table}[!htbp]
\caption{Encode time of the whole database (seconds, $64$ threads) at $\bar{b}$ code bits per dimension.}
\label{tab:vqbench-encode}
\centering
\footnotesize
\setlength{\tabcolsep}{3.5pt}
\resizebox{\linewidth}{!}{%
\begin{tabular}{@{}llrrrrrrrrrr@{}}
\toprule
$\bar{b}$ & \textbf{Method} & \textbf{Llama} & \textbf{Yahoo} & \textbf{ImageNet} & \textbf{COCO} & \textbf{CCNews} & \textbf{ArXiv} & \textbf{HotpotQA} & \textbf{LAION} & \textbf{Landmark} & \textbf{MS MARCO} \\
 & & $d = 128$ & $d = 384$ & $d = 512$ & $d = 768$ & $d = 768$ & $d = 768$ & $d = 640$ & $d = 512$ & $d = 768$ & $d = 1024$ \\
\midrule
$0.5$ & \ezpq & 0.599 & 4.06 & 8.53 & 3.86 & 4.50 & 12.9 & 38.0 & 6.10 & 8.29 & 105 \\
 & OPQ-P & 0.735 & 4.32 & 9.00 & 4.02 & 5.05 & 13.5 & 43.2 & 7.05 & 9.47 & 136 \\
 & OPQ & 0.656 & 4.42 & 9.64 & 4.22 & 5.29 & 15.6 & 52.5 & 7.58 & 10.3 & 145 \\
 & PQ & 0.625 & 3.78 & 8.00 & 3.56 & 4.36 & 11.3 & 39.3 & 6.27 & 8.57 & 104 \\
\midrule
$1$ & \ezpq & 0.995 & 7.94 & 15.8 & 7.24 & 8.23 & 24.6 & 72.3 & 11.5 & 16.8 & 196 \\
 & OPQ-P & 1.25 & 7.85 & 16.4 & 7.50 & 8.59 & 26.6 & 79.3 & 12.1 & 17.2 & 232 \\
 & OPQ & 1.35 & 8.06 & 17.1 & 8.03 & 9.05 & 26.0 & 82.4 & 13.2 & 18.0 & 241 \\
 & PQ & 1.21 & 7.29 & 15.4 & 6.81 & 8.18 & 23.1 & 77.1 & 11.9 & 16.0 & 203 \\
\midrule
$2$ & \ezpq & 1.97 & 15.4 & 31.2 & 10.3 & 15.0 & 41.8 & 141 & 22.1 & 27.2 & 385 \\
 & OPQ-P & 2.56 & 15.2 & 31.3 & 14.7 & 16.3 & 52.2 & 149 & 24.8 & 33.3 & 426 \\
 & OPQ & 2.36 & 15.5 & 31.9 & 14.2 & 17.0 & 48.2 & 150 & 23.7 & 34.0 & 423 \\
 & PQ & 2.31 & 14.5 & 30.2 & 13.5 & 15.6 & 45.8 & 145 & 23.4 & 30.9 & 392 \\
\midrule
$4$ & \ezpq & 3.99 & 29.6 & 55.9 & 19.7 & 28.9 & 79.7 & 263 & 43.2 & 53.5 & 750 \\
 & OPQ-P & 4.75 & 30.5 & 61.2 & 27.2 & 31.5 & 90.2 & 294 & 45.6 & 63.7 & 803 \\
 & OPQ & 4.58 & 30.5 & 61.8 & 28.0 & 32.4 & 88.0 & 293 & 44.6 & 65.5 & 810 \\
 & PQ & 4.57 & 29.5 & 59.3 & 27.0 & 31.1 & 87.5 & 280 & 44.3 & 61.8 & 753 \\
\bottomrule
\end{tabular}}
\end{table}

\begin{table}[!htbp]
\caption{Median time to score one query against its $100$ candidates ($\mu$s) at $\bar{b}$ code bits per dimension.}
\label{tab:vqbench-score}
\centering
\footnotesize
\setlength{\tabcolsep}{3.5pt}
\resizebox{\linewidth}{!}{%
\begin{tabular}{@{}llrrrrrrrrrr@{}}
\toprule
$\bar{b}$ & \textbf{Method} & \textbf{Llama} & \textbf{Yahoo} & \textbf{ImageNet} & \textbf{COCO} & \textbf{CCNews} & \textbf{ArXiv} & \textbf{HotpotQA} & \textbf{LAION} & \textbf{Landmark} & \textbf{MS MARCO} \\
 & & $d = 128$ & $d = 384$ & $d = 512$ & $d = 768$ & $d = 768$ & $d = 768$ & $d = 640$ & $d = 512$ & $d = 768$ & $d = 1024$ \\
\midrule
$0.5$ & \ezpq & 52.9 & 188 & 256 & 375 & 398 & 413 & 539 & 289 & 393 & 1037 \\
 & OPQ-P & 68.1 & 253 & 418 & 647 & 661 & 667 & 862 & 412 & 639 & 2264 \\
 & OPQ & 75.7 & 323 & 532 & 924 & 925 & 1066 & 1139 & 584 & 1005 & 3379 \\
 & PQ & 54.5 & 159 & 208 & 306 & 319 & 308 & 408 & 214 & 315 & 686 \\
\midrule
$1$ & \ezpq & 87.1 & 304 & 460 & 681 & 661 & 714 & 862 & 491 & 711 & 1638 \\
 & OPQ-P & 90.9 & 333 & 553 & 834 & 803 & 1080 & 1061 & 538 & 837 & 2451 \\
 & OPQ & 100 & 407 & 693 & 1080 & 1175 & 1525 & 1350 & 735 & 1208 & 3634 \\
 & PQ & 77.9 & 231 & 345 & 452 & 470 & 612 & 589 & 327 & 468 & 940 \\
\midrule
$2$ & \ezpq & 150 & 519 & 815 & 1196 & 1151 & 1550 & 1446 & 780 & 1402 & 2471 \\
 & OPQ-P & 144 & 493 & 933 & 1141 & 1200 & 1590 & 1330 & 839 & 1445 & 2874 \\
 & OPQ & 161 & 578 & 1117 & 1448 & 1549 & 2092 & 1637 & 1044 & 1689 & 4107 \\
 & PQ & 129 & 393 & 722 & 763 & 889 & 1096 & 928 & 643 & 893 & 1474 \\
\midrule
$4$ & \ezpq & 258 & 980 & 1479 & 2009 & 2327 & 2592 & 2143 & 1577 & 2485 & 3697 \\
 & OPQ-P & 265 & 1029 & 1556 & 2014 & 2470 & 2344 & 1942 & 1620 & 2369 & 3841 \\
 & OPQ & 274 & 1039 & 1715 & 2241 & 2778 & 2739 & 2216 & 1740 & 2776 & 4985 \\
 & PQ & 241 & 850 & 1290 & 1507 & 2106 & 1899 & 1615 & 1289 & 1878 & 2487 \\
\bottomrule
\end{tabular}}
\end{table}
\FloatBarrier


\end{document}